\documentclass[runningheads]{llncs}

\usepackage[mobile]{eccv}

\usepackage{eccvabbrv}

\usepackage{graphicx}
\usepackage{booktabs}

\usepackage[accsupp]{axessibility}  

\usepackage{algorithm}
\usepackage{algpseudocode}
\usepackage{multirow}
\usepackage{pifont}

\definecolor{darkgreen}{RGB}{0, 100, 0}
\definecolor{darkred}{RGB}{139, 0, 0}

\usepackage{pifont}

\usepackage[table]{xcolor}
\definecolor{RowBlueGray}{RGB}{232,240,248}
\definecolor{RowLightBlue}{RGB}{225,238,255}
\definecolor{TrainedRow}{RGB}{250,230,230}

\usepackage{hyperref}

\usepackage{orcidlink}

\begin{document}

\title{VIEW2SPACE: Studying Multi-View Visual Reasoning from Sparse Observations}

\titlerunning{VIEW2SPACE}

\author{Fucai Ke\inst{1} \and
Zhixi Cai\inst{1} \and
Boying Li\inst{1}\and
Long Chen\inst{1}\and 
Beibei Lin\inst{2} \and
Weiqing Wang\inst{1}\and
Pari Delir Haghighi\inst{1}\and 
Gholamreza Haffari\inst{1}\and
Hamid Rezatofighi\inst{1}
}

\authorrunning{F.~Ke et al.}

\institute{Monash University, Australia \and
National University of Singapore\\
\href{https://github.com/pokerme7777/VIEW2SPACE}{Project Page}
}

\maketitle


\begin{abstract}
\vspace{-2em}
Multi-view visual reasoning is essential for intelligent systems that must understand complex environments from sparse and discrete viewpoints, yet existing research has largely focused on single-image or temporally dense video settings. In real-world scenarios, reasoning across views requires integrating partial observations without explicit guidance, while collecting large-scale multi-view data with accurate geometric and semantic annotations remains challenging.
To address this gap, we leverage physically grounded simulation to construct diverse, high-fidelity 3D scenes with precise per-view metadata, enabling scalable data generation that remains transferable to real-world settings. Based on this engine, we introduce VIEW2SPACE, a multi-dimensional benchmark for sparse multi-view reasoning, together with a scalable, disjoint training split supporting millions of grounded question–answer pairs. 
Using this benchmark, a comprehensive evaluation of state-of-the-art vision–language and spatial models reveals that multi-view reasoning remains largely unsolved, with most models performing only marginally above random guessing. We further investigate whether training can bridge this gap. 
Our proposed Grounded Chain-of-Thought with Visual Evidence substantially improves performance under moderate difficulty, and generalizes to real-world data, outperforming existing approaches in cross-dataset evaluation. We further conduct difficulty-aware scaling analyses across model size, data scale, reasoning depth, and visibility constraints, indicating that while geometric perception can benefit from scaling under sufficient visibility, deep compositional reasoning across sparse views remains a fundamental challenge.
\vspace{-0.5em}
\vspace{-1em}
\end{abstract}    
\section{Introduction}
\label{sec:intro}
\vspace{-1em}

\begin{figure*}[t]
\begin{center}
  \includegraphics[width=1.0\linewidth]{pic/teaser.png} 
\end{center}
\vspace{-1.5em}
\caption{A single glance is often insufficient for completing real-world tasks. Humans naturally integrate observations across sparse viewpoints to form a shared spatial understanding. Given heterogeneous views (e.g., robotic dog and drone), one can efficiently align what different agents observe and reason across sparse viewpoints.}
\label{fig:teaser}
\vspace{-1.5em}
\end{figure*}

Visual reasoning in the real world rarely comes from a single glance. When observing complex environments, humans often draw on information gathered from previous viewpoints or actively seek additional perspectives to support judgment and decision making. Even though each individual view may reveal only partial and biased information, people can still infer scene layout, reason about occluded objects, and understand spatial relationships that are never simultaneously visible~\cite{hong20233dmulti-view, gholami2025spatial, yu2025waymoqa, mindcube2025}.

Such multi-view visual reasoning is a foundational capability for intelligent systems operating in complex environments. It enables spatial understanding from sparse, discrete viewpoints, whether acquired by a single observer across locations or collaboratively by multiple observers sharing observations. This setting arises in applications ranging from individual spatial reasoning to collaborative systems such as multi-robot coordination, multi-agent interaction, large-scale multi-camera environments, and autonomous driving, where effective decision making relies on integrating partial visual observations~\cite{ma2025position, ge2021multiagent, gawel2018x}, as illustrated in Figure~\ref{fig:teaser}.

Despite recent progress in vision-language models (VLMs)~\cite{hurst2024gpt4o, bai2025qwenvl25, deitke2025molmo, clark2026molmo2, chen2024internvl}, robust visual reasoning has yet to be consistently achieved even in single-image and densely sampled video settings, which have received the most research attention.
Multi-view visual reasoning is therefore even less understood, despite its importance for comprehensive spatial understanding and downstream tasks (\eg, robotic planning and autonomous driving). A few recent works (\eg, \cite{mindcube2025, deng2025internspatial}) touch on this setting, but they are typically evaluated under non-sparse or human-centric viewpoints, relatively small-scale scenes, and tasks that tend to emphasize perception rather than deeper, multi-hop reasoning across views.
We attribute these limitations to both the difficulty of collecting large-scale real-world multi-view data with accurate per-view annotations and the tendency of existing research to emphasize spatial reconstruction and basic post-hoc relational queries, rather than deeper, multi-hop reasoning over sparse views.

To meaningfully study sparse multi-view reasoning, it is necessary to establish a controlled evaluation setting with large-scale, high-fidelity scenes and objects, together with precise cross-view geometric and semantic metadata. Such a setting should enable systematic variation of viewpoint sparsity, reasoning depth, and task complexity, factors that are difficult to reliably manipulate using real-world data alone, while still supporting transfer to real-world scenarios. We therefore adopt physically grounded simulation as a controllable testbed, designed to minimize the synthetic-to-real gap while preserving geometric and semantic fidelity. This enables scalable generation of high-quality supervision for both training and evaluation.

Within this controlled framework, we build a scalable data engine that generates 3 million QA pairs from 2,000 diverse 3D scenes with precise cross-view geometric and semantic metadata for systematic evaluation and large-scale training. Based on this engine, we introduce VIEW2SPACE, a multi-dimensional benchmark for studying multi-view reasoning under sparse viewpoints.

To assess the current state of multi-view reasoning, we conduct a comprehensive evaluation of state-of-the-art vision–language and spatial models. The results show that multi-view reasoning remains largely unsolved, with most models performing only marginally above random guessing, even on relatively simple reasoning questions. This gap motivates an investigation into whether different training strategies can bridge it, comparing direct-answer, standard CoT, and our proposed Grounded CoT with Visual Evidence. Training consistently improves performance, and grounded CoT further amplifies these gains, achieving up to +52\% mIoU on challenging visual grounding tasks in \textsc{VIEW2SPACE}, measured on a test set covering varying reasoning difficulty levels. Remarkably, the resulting trained model also generalizes strongly to real-world data: when evaluated on the MINDCUBE~\cite{mindcube2025} without further fine-tuning, it surpasses the officially released trained models by over 9\% accuracy.

While these results suggest that multi-view reasoning can be improved under training, a deeper analysis reveals that the problem is far from solved. Scaling across model size, data volume, and task difficulty indicates that although scale improves perceptual performance under sufficient visibility, it exhibits poor scaling efficiency as reasoning complexity increases, suggesting inherent structural limitations in current approaches.
In summary, our contributions are fourfold:
\begin{enumerate}
    \item \textbf{Scalable Data Engine and Benchmark.} We introduce a scalable data engine and \textsc{VIEW2SPACE}, a large-scale benchmark for multi-view spatial reasoning with diverse 3D scenes, geometrically varied viewpoints, rich metadata, graded difficulty, and progressively stricter visual-evidence constraints.

    \item \textbf{Empirical Analysis of Model Behaviour.} We provide a systematic empirical analysis of SOTA models on VIEW2SPACE, revealing performance gaps under increasing evidence constraints, shortcut-based reasoning patterns, and the limitations of explicit thinking in sparse multi-view settings.

    \item \textbf{Visually Grounded Chain-of-Thought Dataset and Method.} We build a step-wise, visually grounded chain-of-thought training dataset and propose Grounded Chain-of-Thought with Visual Evidence, achieving significant gains and strong synthetic-to-real generalization.
    
    \item \textbf{Scaling Analysis of Multi-View Reasoning.} We reveal, through controlled scaling analysis, persistent limitations in deep multi-view reasoning despite increased data and model capacity, suggesting the need for more compositional visual reasoning approaches.
\end{enumerate}

\vspace{-1em}
\section{Literature Review}\vspace{-0.5em}
Recent VLMs such as InternVL~\cite{chen2024internvl}, Molmo~\cite{deitke2025molmo}, Qwen3VL~\cite{Qwen3VL}, and GPTs~\cite{singh2025openaigpt5} have shown promising spatial reasoning abilities through large-scale training. However, their evaluation settings are largely limited to single images, unrelated image collections, or temporally continuous videos, and rarely cover the common scenario of reasoning over sparse, discrete viewpoints of the same scene. Similarly, spatially focused or robotic models (e.g., RoboBrain~\cite{team2025robobrain2, ji2025robobrain}, OpenView~\cite{chen2025openview}, SpatialMLLM~\cite{wu2025spatialmllm}) aim to enhance spatial understanding, but are typically constrained to single-view inputs or rely on explicit spatial supervision.

Existing spatial reasoning benchmarks~\cite{chen2025whatspatial, fu2024blink, lee2025perspective, ma20253dsrbench, qi2025beyond, ramakrishnan2024does, tang2025sparkle, yang2025thinking, zhang2025open3dvqa} predominantly focus on single-view or temporally dense video settings. 
Only a few recent works~\cite{mindcube2025, deng2025internspatial} explore multi-view scenarios. However, these settings are generally not sparse, often adopt human-centric viewpoints, and primarily emphasize perceptual or low-hop reasoning. Moreover, systematic investigation of model shortcut behaviors or guessability under varying evidence constraints remains limited.
In contrast, real-world spatial reasoning requires inferring cross-view relationships directly from visual observations across sparse and heterogeneous viewpoints beyond fixed human perspectives, often involving compositional and multi-hop inference. 
Our work targets this underexplored regime and shows that although perceptual-level multi-view reasoning can be learned with sufficient data, deeper reasoning across discrete and diverse viewpoints remains challenging (Sec.~\ref{experiment_analysis}). Detailed benchmark comparisons are provided in the appendix.

Recent studies have explored incorporating explicit 3D information into LLMs and VLMs to enhance spatial reasoning. One line of work directly introduces structured 3D inputs, such as point clouds~\cite{chen2024ll3da, huang2023embodied} and depth maps~\cite{zhu2024llava3d}, by projecting 3D geometric features into a shared vision–language embedding space. Building on this paradigm, methods such as 3D-R1~\cite{huang20253dr1} and SpatialThinker~\cite{batra2025spatialthinker} further employ explicit chain-of-thought supervision and reinforcement learning with spatial rewards to improve multi-step 3D reasoning, often coupled with view selection or scene graph representations.

However, many of these methods~\cite{chen2024ll3da, huang2023embodied, zhu2024llava3d, huang20253dr1, batra2025spatialthinker} either rely on input modalities beyond RGB that are uncommon in everyday settings, or primarily focus on 3D reconstruction rather than semantic understanding. As a result, their applicability to semantic tasks such as visual question answering or visual grounding under realistic RGB-only conditions remains limited. In contrast, we focus on multi-view spatial reasoning directly from readily available RGB observations.

\vspace{-1em}
\section{Multi-view Data Engine}
\label{sec:multiview_data_engine}
\vspace{-0.5em}

\begin{figure*}[t]
\begin{center}
  \includegraphics[width=1.0\linewidth]{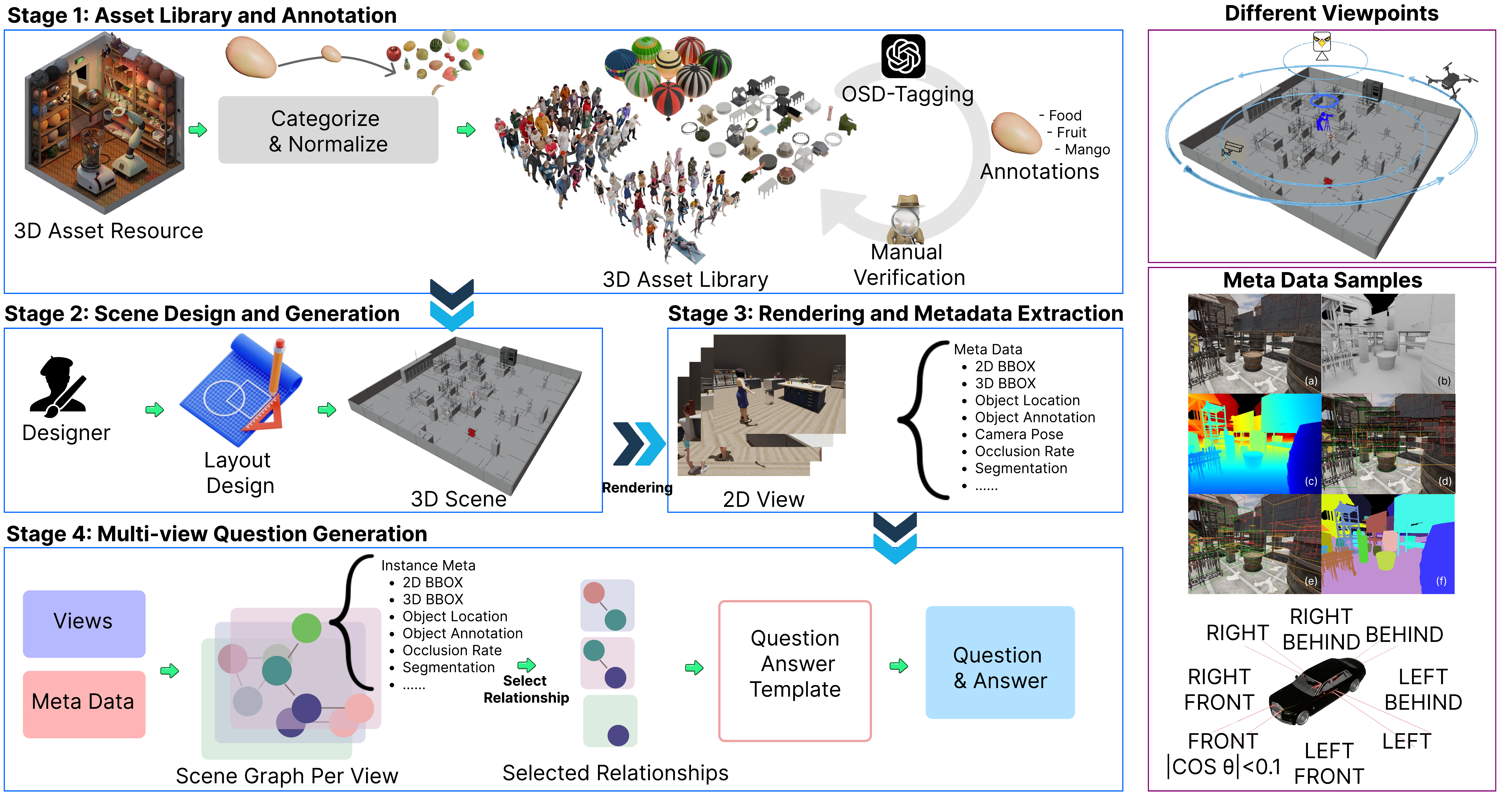} 
\end{center}
\vspace{-1.5em}
\caption{Overview of the multi-view data engine. Stage 1: Assets are collected, categorized, and size-normalized, and annotated using OSD-Tag with verification.
Stage 2: Given theme configurations defining asset categories and spatial relations, scenes are synthesized by sampling layouts under compositional constraints.
Stage 3: Multi-view images are rendered together with physics-based metadata.
Stage 4: Spatial relations are selected to automatically generate grounded question-answer pairs.}
\label{fig:pipeline}
\vspace{-1em}
\end{figure*}

Our multi-view data engine is designed to produce high-fidelity, physically grounded 3D scenes with precise and verifiable cross-view annotations. It constructs diverse environments from over \emph{1,000 high-quality photorealistic assets} with multi-level semantic labels, supporting flexible composition across more than \emph{40 everyday themes}. Built on a physically grounded 3D engine~\cite{community2018blender}, it enables scalable scene generation with geometry-consistent supervision across views. In the following, we describe the overall design and generation pipeline (Figure~\ref{fig:pipeline}).

\textbf{3D Asset Library and Automatic Annotation Pipeline.}
To ensure high realism and physical fidelity, we manually curate a large collection of 3D assets, prioritizing high-quality, scan-based objects that closely reflect real-world geometry and appearance. Assets expected to co-occur within the same scene theme are organized into category-level clusters to maintain semantic coherence during scene composition and annotation. To guarantee metric accuracy and spatial validity, asset scales are normalized to real-world proportions. This calibration integrates prior knowledge from LLMs with manual verification, ensuring object dimensions remain physically consistent. All assets are further assigned semantic categories based on functional role and appearance, forming the foundation for structured scene generation and hierarchical annotation.

\textbf{Overview-driven Semantic Differentiation and Tagging (OSD-Tag).}
Given the need for a scalable and largely automated yet reliable annotation pipeline, we introduce Overview-driven Semantic Differentiation and Tagging. Using aggregated asset previews within each category as an overview, OSD-Tag leverages a VLM (\ie, GPT-4o~\cite{hurst2024gpt4o}) to identify key visual and functional differences and construct a hierarchical semantic tag library. These tags are then automatically assigned to each asset, followed by \emph{manual verification} to ensure correctness. The resulting hierarchical annotations support controlled scene composition, question generation, and diagnostic analysis. Implementation details are provided in the appendix.

\textbf{Thematic Scene Design and Layout Generation.} To ensure realistic scene composition while maintaining layout diversity, each scene theme is manually designed with explicit semantic and spatial constraints. Designers define theme-level configurations that govern object categories, placements, and inter-category relationships, enabling coherent yet varied scene generation. These constraints are encoded in theme-specific configuration files and used to sample valid layouts with controlled randomness. Scene generation is implemented in Blender~\cite{community2018blender}, enabling precise geometric control and adjustable density, thereby producing diverse but structurally consistent scenes within each theme.

\textbf{Multi-View Rendering and Metadata Extraction.}
To support the diverse requirements of benchmarking, diagnostic analysis, and model training, we design a set of four canonical viewpoint configurations for multi-view rendering, including drone views, bird-eye views, human-like egocentric views, and fixed surveillance camera views. Within each viewpoint category, users can flexibly control camera placement and orientation (\eg, viewing the scene from the center or from peripheral locations, and from different azimuthal directions), as well as camera coverage and optics, such as the proportion of the scene captured and the use of wide-angle versus narrow field-of-view settings. These viewpoints capture scenes from complementary perspectives and introduce systematic variation in viewpoint geometry and visibility.

Using the underlying physics and rendering engines, the system automatically computes rich view-dependent metadata, including object bounding boxes, occlusion ratios, and camera-relative visibility statistics (see figure in the appendix).
We further record precise geometric information for each object, such as 3D pose, orientation, and spatial relationships, which are stored alongside rendered images. This comprehensive metadata enables fine-grained evaluation, controlled diagnostic studies, and the generation of structured supervision for multi-view visual reasoning.

\textbf{Grounded Multi-View Question and Supervision Generation.} 
We construct question templates spanning multiple reasoning difficulty levels to promote diversity while maintaining answer reliability. Answers are then deterministically derived through rule-based mapping from scene geometry, avoiding heuristic or generative inference. Using controllable 3D scenes and engine-provided ground-truth metadata, including object identities, 3D poses, camera parameters, view-dependent bounding boxes, and occlusion statistics, we compute spatial relations in both object-centric and camera-centric coordinate frames. This process produces strictly grounded multi-view question–answer pairs and structured supervision such as instance-level grounding, explicit inter-view relations, and geometry-derived reasoning traces. Additional implementation details are provided in the appendix.

\vspace{-1em}
\begin{figure*}[t]
\begin{center}
  \includegraphics[width=1.0\linewidth]{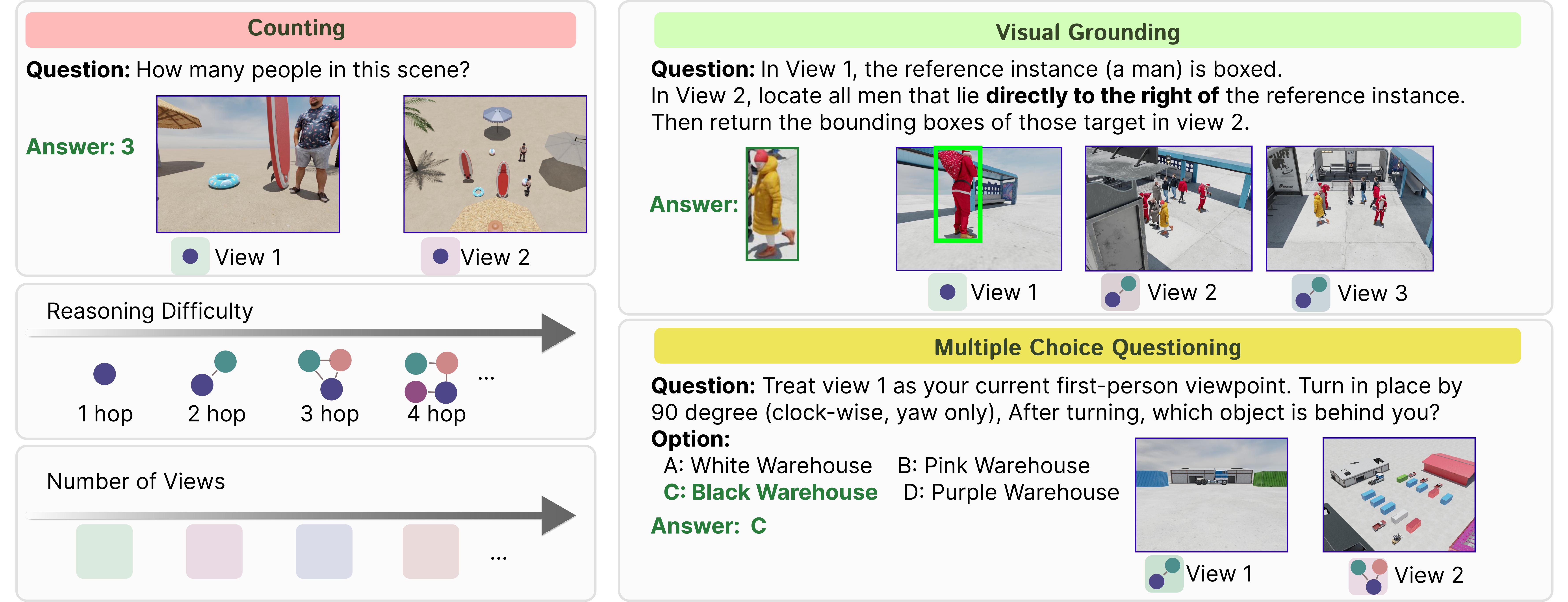} 
\end{center}
\vspace{-1.5em}
\caption{Overview of VIEW2SPACE task design.
VIEW2SPACE comprises three task types: MCQ, counting, and visual grounding detection. Reasoning difficulty ranges from one-hop perceptual queries to multi-hop cross-view spatial reasoning.}
\label{fig:benchmark_sample}
\vspace{-1.5em}
\end{figure*}

\section{VIEW2SPACE Benchmark and Evaluation}\vspace{-0.5em}
\label{experiment_analysis}
\textsc{VIEW2SPACE} is designed to study multi-view spatial reasoning under sparse and geometrically diverse viewpoints, where no single view is sufficient to resolve the query. Beyond simply providing multiple images, the benchmark covers a graded spectrum of reasoning difficulty, from basic cross-view alignment to high-dimensional, multi-hop compositional reasoning, and includes task formats with increasingly strict visual-evidence constraints (MCQ, counting, and detection).

Empirically, most models perform only marginally above random guessing even on multiple-choice questions, which are the most permissive and tolerant to informed guessing. When explicit visual grounding is required, performance drops further, with detection tasks yielding near-zero results. This pattern indicates that current models struggle to integrate information across sparse viewpoints and fail under stricter grounding constraints.

\vspace{-1em}
\subsection{VIEW2SPACE Benchmark}

\textbf{Overview}
We introduce \textsc{VIEW2SPACE}, a benchmark for evaluating multi-view spatial reasoning under sparse viewpoints. The benchmark tests whether models can integrate incomplete cross-view observations into a coherent spatial representation as shown in Figure~\ref{fig:benchmark_sample}. \textsc{VIEW2SPACE} comprises groups of images capturing the same static scene from different viewpoints, paired with reasoning questions whose answers require cross-view inference. In contrast to prior benchmarks~\cite{liu2023visualspatialreasoning, cheng2024spatialrgpt, chen2024spatialvlm, yang2025vsibench, pothiraj2025capture, mindcube2025}, which either focus on single-image understanding or continuous video streams, or involve human-centric viewpoints with limited layout diversity and predominantly low-hop perceptual queries, \textsc{VIEW2SPACE} is designed to evaluate deep compositional reasoning across sparse and geometrically diverse viewpoints.

\textbf{Task Types and Evaluation Metric.}
To our knowledge, \textsc{VIEW2SPACE} is the first benchmark that systematically spans tasks from highly permissive answer spaces to strictly constrained grounding-based detection.
It includes three task types: multiple-choice questioning (MCQ), counting, and visual grounding detection, that progressively restrict the feasible answer space and reduce the likelihood of correct predictions obtained through guessing or unsupported reasoning. Rather than differing only in output structure, these tasks reflect increasing resistance to correct answers obtained through linguistic bias or hallucinated reasoning. 
MCQ is the most permissive, allowing correctness via partial evidence or informed guessing, and is evaluated using accuracy (ACC). Counting requires preliminary grounding of relevant instances before aggregation, narrowing plausible guesses, and is evaluated using ACC and mean absolute error (MAE). Visual grounding detection is the most restrictive, demanding explicit localization of target objects and is evaluated using mIoU and F1-score (IoU $>$ 50\%), making unsupported yet correct answers least likely.

\textbf{Reasoning Difficulty.}
Within each task type, questions are organized along a graded reasoning difficulty spectrum, ranging from basic perceptual alignment across views to compositional, multi-hop spatial reasoning. Difficulty is defined by the minimal number of relational hops required in a per-view object–relation graph, where objects are nodes and spatial relations are edges. A single perceptual grounding step counts as one hop, and complexity further increases with the logarithm of the number of key objects involved. Detailed reasoning difficulty settings are provided in the Appendix.

\textbf{Object Visibility Difficulty.}\label{def_object_vis_define} We define it based on the proportion of an object’s projected area on the image plane that is not directly visible to the camera. Using the underlying Blender rendering engine, we perform ray casting from object surface points toward the camera to determine whether each point is directly observable. The fraction of rays that fail to reach the camera due to occlusion is taken as the visibility difficulty score.
For each task, we define \textit{key-object visibility difficulty} as the average visibility difficulty of all task-relevant objects aggregated across views.

\textbf{Viewpoint Configurations.}
VIEW2SPACE systematically varies viewpoint pairings to probe spatial reasoning under heterogeneous perspectives, as shown in the \emph{Different Viewpoints} block in Figure~\ref{fig:pipeline}. These pairings include combinations of human-like egocentric views, drone views, bird’s-eye views, and fixed surveillance camera views. By combining viewpoints with distinct geometric properties, the benchmark creates controlled variations in viewpoint geometry and visibility, enabling evaluation of reasoning under diverse cross-view conditions.

\textbf{Statistics of VIEW2SPACE.}
We introduce VIEW2SPACE-v1, a curated evaluation split comprising $3{,}591$ questions (1,400 MCQ, 591 Counting, and 1,600 Detection), with balanced answer distributions across 40 themes and 1,096 primary 3D assets spanning graded difficulty levels. As sparse multi-view reasoning is already challenging for current VLMs, even lower-difficulty cases remain non-trivial; thus, we maintain a graded distribution rather than concentrating on extreme difficulty.
Although VIEW2SPACE scales to \textbf{3 million samples}, large-scale evaluation, particularly on closed-source models, incurs substantial computational and API costs. To balance evaluation cost (approximately \$100 per closed-source model) and reproducibility, we benchmark the 3k-sample VIEW2SPACE-v1 in this work. Detailed statistics are provided in the Appendix.

\textbf{A Scalable Training Split for VIEW2SPACE.}
We construct a scalable training dataset with a size of 300K with scenes disjoint from the evaluation split, supporting expansion to millions of question–answer pairs. Each instance is derived from engine-level geometric and physical ground truth, including bounding boxes, occlusion ratios, and precise cross-view spatial relations. Via a symbolic pipeline, we generate deterministic, visually grounded chain-of-thought supervision with verifiable intermediate steps, eliminating the stochastic noise and inconsistency of MLLMs generated reasoning traces and yielding clean, geometry-consistent training signals. 
Additional details are provided in the appendix.

\begin{table}[!t]

\caption{Comparison on \textsc{VIEW2SPACE-v1}}
\vspace{-1em}
\label{tab:evaluation}
\centering
\scalebox{0.58}{
\begin{tabular}{lc|cc|cc}
\toprule[0.3mm]
\rowcolor[gray]{0.8} Method & \multicolumn{1}{c|}{Multiple Choice Answering} & \multicolumn{2}{c|}{Visual Counting } & \multicolumn{2}{c}{Visual Grounding} \\
\rowcolor[gray]{0.8}& ACC (\%) & MAE & ACC (\%) & mIoU (\%) & F1-score (\%)\\

\rowcolor[gray]{0.97} Baseline & & & & &\\
Random (chance)         & 28.59 & 11.37 & 3.39 & 0.22 & 0.0\\
Random (frequency)      & 28.59 & 2.46 & 13.05 & 2.01 & 0.27 \\
\hline
\rowcolor[gray]{0.97} Open-source MLLMs  & & & & &\\

Internvl2.5-8B~\cite{chen2024internvl25} & 30.93 & 2.04 & 22.50 & 0.79 & 0.0\\
Mantis-8B (SigLip)~\cite{jiangmantis} & 30.07 & 2.94 & 17.60 & 1.24 & 0.23 \\

DeepSeek-VL2-Small~\cite{lu2024deepseekvl} & 29.43 & 2.20 & 24.37 & 2.86 & 0.19 \\

Gemma-3-12B-it~\cite{team2025gemma3} & 31.14 & 2.08 & 24.53 & 2.49 & 0.24 \\

Idefics2-8B~\cite{laurenccon2024idefics2} & 26.64 & 3.92 & 11.34 & 0.39 & 0.0 \\

Qwen2.5-VL-7B~\cite{bai2025qwenvl25}  & 30.29 & 2.70 & 13.03 & 2.68 & 0.32 \\

Qwen3-VL-2B~\cite{Qwen3VL}    & 31.43 & 2.63 & 16.34 & 11.20 & 5.07 \\

Qwen3-VL-4B~\cite{Qwen3VL}    & 35.19 & 2.26 & 21.15 & \cellcolor[rgb]{1.00,0.97,0.80}\underline{16.53} & \cellcolor[rgb]{1.00,0.97,0.80}\underline{18.72}\\

Qwen3-VL-8B~\cite{Qwen3VL}    & 37.41 & 2.10 & 24.92 & 10.05 & 7.37 \\
\hline
\rowcolor[gray]{0.97} Open-source Spatial Models  & & & & &\\
Molmo2-4B~\cite{clark2026molmo2}   & 31.36 & 2.21 & 22.17 & 1.17 & 0.13 \\
Molmo2-8B~\cite{clark2026molmo2}   & 34.57 & 5.93 & 21.66 & 1.51 & 0.0  \\
RoBobrain2.0~\cite{team2025robobrain2}& 30.79 & 2.22 & 21.49 & 0.0 & 0.0\\
Spatial-MLLM~\cite{wu2025spatialmllm}& 27.93 & 2.10 & 9.98 & 0.0 & 0.0\\
Openview~\cite{chen2025openview} & 30.72 & 2.57 & 19.80 & 1.08 & 0.30\\
MINDCUBE-Aug~\cite{mindcube2025}& 29.00 & 4.52 & 18.10 & 0.0 & 0.0 \\
MINDCUBE-Plain~\cite{mindcube2025}& 30.21 & 4.77 & 19.12 & 0.12 & 0.0 \\
\hline

\hline
\rowcolor[gray]{0.97} Closed-source MLLMs  & & & & & \\
GPT-4o~\cite{hurst2024gpt4o} & 33.93 & 1.51 & 29.03 & 3.60& 0.56\\
GPT-5-nano~\cite{singh2025openaigpt5}  & 33.36 & 1.73 & 24.28 & 2.89 & 0.87 \\
GPT-5-mini~\cite{singh2025openaigpt5}  & 49.00 & \cellcolor[rgb]{1.00,0.97,0.80}\underline{1.20} & \cellcolor[rgb]{1.00,0.97,0.80}\underline{38.71} & 7.42 & 2.77 \\

GPT-5~\cite{singh2025openaigpt5}  & \cellcolor[rgb]{1.00,0.97,0.80}\underline{59.86} & 1.31 & 38.10 & 8.18 & 3.43 \\

GPT-5.2~\cite{openai2025gpt52}  & 37.07 & 1.28 & 36.50 & 8.74 & 2.84 \\

\hline
\rowcolor{TrainedRow} Fine-tuned Models (Qwen3VL-4B) & & & & & \\
Ours (Instruct-tuning) 
& 36.50
& 0.71
& 50.08
& 47.58
& 50.07\\

Ours (Grounded CoT, Direct Answer at Test)
& 33.95
& 10.75
& 29.10
& 16.35
& 11.50\\

Ours (CoT) 
& 59.57
& 0.68
& 51.43
& 49.77
& 52.34 \\

Ours (Grounded CoT) 
& \cellcolor[rgb]{0.85,1.00,0.85}\textbf{64.93} 
& \cellcolor[rgb]{0.85,1.00,0.85}\textbf{0.58} 
& \cellcolor[rgb]{0.85,1.00,0.85}\textbf{54.99}
& \cellcolor[rgb]{0.85,1.00,0.85}\textbf{69.34} 
& \cellcolor[rgb]{0.85,1.00,0.85}\textbf{70.92} \\

\textbf{\textcolor{orange}{$\Delta$(abs)} / \textcolor{blue}{$\Delta$(\%)}} 
& $\uparrow$ \textbf{\textcolor{orange}{5.07} / \textcolor{blue}{8.47}}
& $\downarrow$ \textbf{\textcolor{orange}{0.62} / \textcolor{blue}{51.67}}
&$\uparrow$ \textbf{\textcolor{orange}{16.28} / \textcolor{blue}{42.06}}
& $\uparrow$ \textbf{\textcolor{orange}{52.81} / \textcolor{blue}{319.48}} 
& $\uparrow$ \textbf{\textcolor{orange}{52.2} / \textcolor{blue}{278.85}} \\

\bottomrule[0.3mm]
\end{tabular}}
\vspace{-2em}
\end{table}

\vspace{-0.5em}
\subsection{Evaluation on VIEW2SPACE}\label{think_nocot_inmain}\vspace{-0.5em}
We evaluate models on \textsc{VIEW2SPACE-v1}, a difficulty-balanced split with a relatively higher proportion of lower-difficulty cases. Since sparse multi-view reasoning remains underexplored and most models struggle even on basic multi-view and grounding tasks, this design avoids over-concentrating evaluation on only the hardest instances. The evaluated models include open-source MLLMs~\cite{chen2024internvl25, jiangmantis, lu2024deepseekvl, team2025gemma3, laurenccon2024idefics2, Qwen3VL, bai2025qwenvl25}, specialized spatial reasoning models~\cite{clark2026molmo2, ji2025robobrain, wu2025spatialmllm, chen2025openview, mindcube2025} with explicit fine-tuning on spatial reasoning dataset, as well as closed-source MLLMs~\cite{hurst2024gpt4o, openai2025gpt52, singh2025openaigpt5}. An overview of all evaluated models is provided in Table~\ref{tab:evaluation}. 
As reference, we include two baselines: Random (chance), which samples uniformly, and Random (frequency), which follows the empirical answer distribution, estimating lower-bound performance under different prior knowledge assumptions.

\textbf{Performance Across Tasks with Shrinking Answer Space.} To our knowledge, VIEW2SPACE is the first benchmark that spans tasks ranging from those more tolerant to hallucinated reasoning to those least tolerant, progressively enforcing stricter visual-evidence constraints and tighter solution spaces.

Model performance consistently declines as visual-evidence constraints tighten, closely tracking the reduction in answer space from left to right in Table~\ref{tab:evaluation}. Under strict grounding constraints, most models perform near random baselines, indicating limited grounding capability and reflecting the current capability profile of VLMs, which remain stronger in text-centric generation and selection than in precise localization. Importantly, this does not imply that grounding is unattainable. Recent efforts increasingly incorporate grounding-oriented training into VLMs (\ie, \cite{chen2024internvl25, Qwen3VL, clark2026molmo2}). Notably, Qwen3VL~\cite{Qwen3VL} performs measurably above baseline despite low absolute accuracy; consistent with its reported grounding-focused training, this suggests that grounding robustness is achievable with explicit supervision. The GPTs~\cite{hurst2024gpt4o,singh2025openaigpt5,openai2025gpt52} likewise show steady improvements in grounding performance across versions, aligning with their reported enhancements. In contrast, although Molmo2~\cite{clark2026molmo2} also emphasizes grounding, its performance remains limited under sparse multi-view constraints, indicating that grounding learned in conventional settings may not readily transfer to multi-view reasoning.

\textbf{Comparative Performance of State-of-the-Art Models.}
Table~\ref{tab:evaluation} shows that most models retain substantial headroom even on the most hallucination-tolerant MCQ tasks. We do not interpret this as an inability to perform multi-view reasoning per se, but rather as a limited transfer of spatial skills learned in conventional settings to sparse multi-view scenarios without dedicated training.

Closed-source GPTs~\cite{hurst2024gpt4o, singh2025openaigpt5} exhibit relatively stable performance; for instance, GPT-5~\cite{singh2025openaigpt5} achieves 59.86\% accuracy on MCQ. This may relate to larger model scale and exposure to diverse perceptual and reasoning data, potentially including 3D-derived or perception-heavy VQA-style supervision. Interestingly, GPT-5.2~\cite{openai2025gpt52} does not outperform GPT-5~\cite{singh2025openaigpt5} or GPT-5-mini, and performs closer to GPT-4o~\cite{hurst2024gpt4o} on MCQ. While GPT-5.2~\cite{openai2025gpt52} reports improvements in grounding and an increased emphasis on coding, these enhancements do not translate into clear gains in our multi-view setting, suggesting that fine-tuning objectives may not strongly target sparse multi-view spatial alignment.

Open-source spatial models do not show consistent advantages over general open-source MLLMs, and generally underperform the Qwen3-VL series~\cite{Qwen3VL} as shown in Table~\ref{tab:evaluation}. This suggests that spatial capabilities developed through robotic spatial training or static single-view supervision may have limited transfer to discrete multi-view reasoning. While single-view spatial understanding remains fundamental, multi-view reasoning additionally requires cross-view alignment and geometric calibration, which likely accounts for the observed gap. Finally, as reasoning difficulty progressively increases within the dataset, overall performance trends also reflect the structured escalation of task complexity.

\textbf{Thinking vs. Non-Thinking Inference.}
In the main experiments, models were prompted without restricting reasoning style. We additionally compare thinking (CoT) and direct-answer inference for open-source MLLMs (Appendix). As grounding performance is near zero for most of these models, we focus on counting and MCQ tasks.
Explicit thinking does not improve performance and may degrade accuracy. Manual inspection suggests that in sparse multi-view settings, direct answers may exploit shortcut cues. In contrast, enforced thinking often produces repetitive or weakly grounded reasoning, extending the reasoning trace without effectively integrating cross-view evidence and ultimately leading to incorrect conclusions. This indicates that explicit reasoning without sufficient multi-view training may introduce instability rather than gains.

These findings raise a central question: \textbf{can sufficient data and appropriate training enable models to acquire spatial reasoning for sparse multi-view settings?} We investigate this by training on VIEW2SPACE.

\section{From Data to Multi-View Reasoning: What Can Training Achieve?}
In this section, we investigate how different training strategies affect multi-view visual reasoning. We compare standard instruction tuning, language-only chain-of-thought training, and our proposed Grounded Chain-of-Thought with Visual Evidence to understand the role of explicit visual grounding. We further examine synthetic-to-real generalization and analyze how performance scales with model size, data volume, reasoning difficulty, and visibility constraints. Together, these experiments aim to clarify what training can and cannot achieve for sparse multi-view reasoning.

\subsection{Supervised Fine-Tuning for Multi-View Reasoning}
In this subsection, we study how supervised fine-tuning enables multi-view visual reasoning, examining its effectiveness under different supervision strategies.

\textbf{Training Setup.}
We adopt Qwen3-VL-4B-Instruct~\cite{Qwen3VL} as the base model and perform full-parameter supervised fine-tuning under three strategies: instruction tuning, standard CoT, and our proposed grounded CoT with visual evidence. Training samples at the 300K scale are generated by our data engine. Additional implementation details are provided in the appendix.

\textbf{Limited Gains from Instruction Tuning.}
We evaluate the effectiveness of monolithic instruction tuning for multi-view visual reasoning as shown in Table~\ref{tab:evaluation}. While instruction tuning consistently outperforms random guessing on VIEW2SPACE-v1, the overall gains remain limited. In particular, performance on MCQ tasks (36.50\% ACC) falls well below that of GPT-5~\cite{singh2025openaigpt5} (59.86\% ACC), indicating that instruction tuning alone is insufficient for learning deeper cross-view reasoning patterns.

Instruction tuning yields relatively stronger improvements on counting and visual grounding tasks, where supervision is directly anchored to visual evidence. Consistent with our analysis in Sec.~\ref{think_nocot_inmain}, we attribute this to task familiarity: these tasks align more closely with the types of visually grounded supervision that the community has increasingly emphasized, and instruction tuning provides missing multi-view training signals that help models develop initial competence in this unfamiliar setting. 
These results suggest that instruction tuning alone is insufficient for learning structured cross-view relational reasoning, motivating the exploration of more explicitly grounded training strategies in the following subsection.

\textbf{CoT Improves MCQ but Not Grounding-Intensive Tasks.}
As shown in Table~\ref{tab:evaluation}, standard CoT reasoning yields substantial improvements on MCQ tasks, increasing accuracy from 36.50\% under instruction tuning to 59.57\%, approaching the performance of GPT-5~\cite{singh2025openaigpt5} on the same setting. However, compared to instruction tuning, improvements on counting and visual grounding tasks remain limited, despite their stronger reliance on localization ability. This suggests that language-driven reasoning alone may not sufficiently compensate for missing visual grounding capability, indicating a bottleneck when grounding is required.

\textbf{Grounded Chain-of-Thought Enables Broad Improvements.}
Beyond standard instruction tuning, we propose a grounded CoT training strategy that explicitly ties intermediate reasoning steps to visual evidence (explicitly output key object bounding box in the corresponding view). Instead of relying on free-form textual reasoning, the model is encouraged to reference visually grounded entities and spatial relations when forming intermediate conclusions.

This design reduces ungrounded linguistic reasoning and promotes reasoning anchored in observable visual cues across multiple views. Empirically, grounded CoT leads to consistent and substantial improvements across tasks. On MCQ, it achieves over 77\% relative accuracy improvement compared to standard instruction tuning. More notably, on hallucination-sensitive visual grounding tasks, grounded CoT yields over 300\% relative improvement in detection mIoU, indicating significantly stronger evidence-grounded reasoning. These results demonstrate that explicitly grounding intermediate reasoning steps in visual evidence is critical for robust multi-view reasoning beyond surface-level pattern matching.

These results suggest that explicitly incorporating visual evidence into intermediate reasoning steps is essential for robust multi-view reasoning, especially in settings where correct answers cannot be obtained through linguistic shortcuts.

\begin{table}[!t]
\caption{
Comparison on MINDCUBE-Tiny. The second column indicates whether models are fine-tuned on MINDCUBE (w Tuning vs. w/o Tuning).
}
\vspace{-1em}
\label{tab_syn_to_real}
\centering
\scalebox{0.7}{
\begin{tabular}{lcc|ccc}
\toprule[0.3mm]
\rowcolor[gray]{0.8} Method & Tuning on MINDCUBE  & Overall & Rotation & Among & Around\\

\rowcolor[gray]{0.97} Baseline & & & &\\
Random (chance)      & - & 32.35 & 36.36 & 32.29 & 30.66 \\
Random (frequency)   & - & 33.02 & 38.30 & 32.66 & 35.79 \\
\hline
\rowcolor{RowBlueGray}\multicolumn{6}{l}{Open-source MLLMs }\\

Internvl2.5-8B~\cite{chen2024internvl25}  & w/o Tuning  & \cellcolor[gray]{0.9} 18.68 & 36.45 & 47.11 & 26.91 \\

Molmo2-4B~\cite{clark2026molmo2}  & w/o Tuning  & 37.33 & 36.00 & 38.83 & 35.75 \\
Molmo2-8B~\cite{clark2026molmo2}  & w/o Tuning  & 34.33 & 27.50 & 36.50 & 34.50 \\

Mantis-8B (SigLip)  & w/o Tuning  & 41.05 & 37.65 & 40.23 & 50.99\\
DeepSeek-VL2-Small  & w/o Tuning  & 47.62 & 37.00 & 50.38 & 26.91\\
Gemma-3-12B-it  & w/o Tuning  & 46.67 & 38.39 & 48.38 & 34.63\\
Idefics2-8B  & w/o Tuning  & 35.86 & 35.15 & 35.94 & 35.49\\
Qwen2.5-VL-7B~\cite{bai2025qwenvl25}  & w/o Tuning  & 29.26 & 38.76 & 29.50 & 21.35 \\

Qwen3-VL-2B~\cite{Qwen3VL} & w/o Tuning  & 32.58 & 26.50 & 36.33 & 30.00 \\
Qwen3-VL-4B~\cite{Qwen3VL} & w/o Tuning  & 36.67 & 41.50 & 34.67 & 37.25 \\
Qwen3-VL-8B~\cite{Qwen3VL} & w/o Tuning  & 37.42 & 46.00 & 35.83 & 35.50 \\
RoBobrain~\cite{ji2025robobrain} & w/o Tuning  & 37.38 & 35.80 & 38.28 & 29.53 \\
Spatial-MLLM~\cite{wu2025spatialmllm} & w/o Tuning  & 32.06 & 38.39 & 20.92& 32.82 \\
Openview~\cite{chen2025openview} & w/o Tuning  & 40.08 & 24.50 & 45.50 & 39.75 \\

\rowcolor{RowBlueGray}\multicolumn{6}{l}{Closed-source MLLMs} \\
Claude-4-Sonnet & w/o Tuning  & 44.75 & 48.42 & 44.21 & 47.62 \\
GPT-4o          & w/o Tuning  & 38.81 & 32.65 & 40.17 & 29.16 \\
GPT-5-nano      & w/o Tuning  & 38.58 & 44.00 & 38.67 & 35.75 \\
GPT-5-mini      & w/o Tuning  & 46.64 & 89.00 & 44.67 & 51.75 \\

\rowcolor{TrainedRow}\multicolumn{6}{l}{MINDCUBE-Checkpoints}\\
MINDCUBE-Aug~\cite{mindcube2025}  & w Tuning  & 55.24 & 49.50 & 52.50 & 66.40 \\
MINDCUBE-Plain~\cite{mindcube2025}& w Tuning  & \cellcolor[rgb]{1.00,0.97,0.80} \underline{60.76} & 47.50 & 62.33 & 67.60 \\
\hline
\rowcolor{RowBlueGray}\multicolumn{6}{l}{VIEW2SPACE-Trained Models}\\
Ours 2B (Grounded-CoT)  & w/o Tuning  & 59.92 & 49.00 & 56.50 & 70.50 \\
Ours 4B (Grounded-CoT)  & w/o Tuning   & \cellcolor[rgb]{0.85,1.00,0.85} \textbf{70.00} & 71.50 & 66.33 & 74.75 \\
\textbf{\textcolor{orange}{$\Delta$(abs)} / \textcolor{blue}{$\Delta$(\%)}} &  & $\uparrow$ \textbf{\textcolor{orange}{9.24} / \textcolor{blue}{15.21}} \\

\bottomrule[0.3mm]
\end{tabular}}
\vspace{-2em}
\end{table}

\subsection{Synthetic-to-Real Generalization}
Beyond in-domain evaluation, we assess transfer to real-world data. We evaluate on MindCube-Tiny~\cite{mindcube2025}, a real-world multi-view dataset, following the official evaluation protocol. Compared to our benchmark, MindCube frequently encodes inter-view relations through language descriptions, employs a relatively limited set of viewpoint configurations, and involves simpler object-relation structures. Moreover, it contains only MCQ without grounding-based evaluation. Results are shown in Table~\ref{tab_syn_to_real}.

To assess cross-domain generalization, we compare models trained solely on VIEW2SPACE with models fine-tuned directly on the MindCube training set. Despite not being trained on any MindCube data, VIEW2SPACE-trained models outperform the official MindCube checkpoints by 9.24\% absolute accuracy and 15.21\% relative gain. These results demonstrate that the supervision provided by VIEW2SPACE generalizes effectively across domains, even when transferring from synthetic sparse-view settings to real-world data.

\subsection{Limits of Scaling in Multi-View Reasoning}

\begin{figure*}[t]
\begin{center}
  \includegraphics[width=1.0\linewidth]{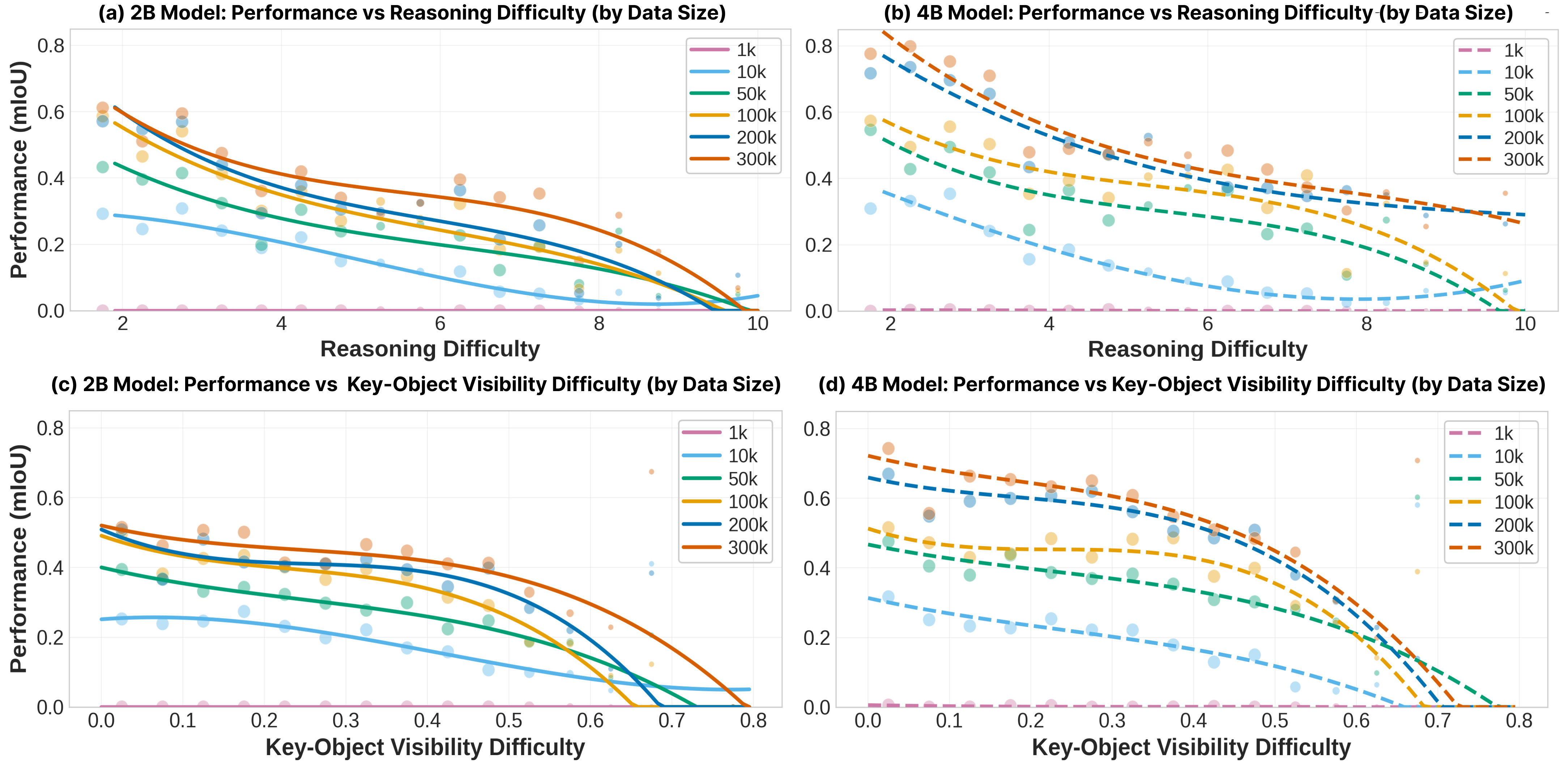} 
\end{center}
\vspace{-1.5em}
\caption[Scaling behaviour on the detection task.]{Scaling Behaviour across Reasoning and Key-Object Visibility Difficulty (see Sec.~\ref{def_object_vis_define}). Each subplot plots mIoU versus task difficulty. Each curve corresponds to a specific dataset size, fitted over the averaged performance at each difficulty level (aggregated across multiple samples and weighted accordingly). The dense empirical dot plots and detailed results are provided in the appendix.}
\label{fig:performance_analysis}
\vspace{-2em}
\end{figure*}

Despite strong overall performance and encouraging synthetic-to-real transfer, multi-view reasoning remains far from solved. VIEW2SPACE-v1 spans a range of difficulty from perceptual alignment to multi-hop reasoning; given the novelty of sparse multi-view tasks, a substantial portion remains perceptual or low-hop. As a result, aggregate metrics average across difficulty levels and can obscure model performance on genuinely complex cases. 
To probe the limits of training and assess the effects of dataset and model scale on multi-view reasoning, we conduct controlled experiments on a balanced 3K-sample analysis set. We fine-tune Qwen3-VL-2B/4B~\cite{Qwen3VL} with increasing training data (1K–300K) and evaluate performance across reasoning and key-object visibility difficulty (Figure~\ref{fig:performance_analysis}). To isolate factors, reasoning analysis restricts visibility difficulty below 0.5, while visibility analysis restricts reasoning difficulty below 6. Additional details are provided in the appendix.

Our analysis shows that while scaling dataset and model size improves multi-view perceptual capability and yields strong performance under sufficient visibility, it exhibits poor scaling efficiency as reasoning complexity increases. As illustrated in Figure~\ref{fig:performance_analysis}, training improvements span roughly 70\% of the visibility spectrum shown, whereas diminishing returns already emerge within only a tiny portion of the effectively unbounded reasoning complexity spectrum. This suggests that scaling may raise performance ceilings without removing underlying structural constraints.

\textbf{Visibility difficulty reflects a learnable perceptual regime with an observability limit.} Performance declines gradually across a broad range of increasing visibility difficulty, indicating that current methods can learn effective geometric and perceptual representations when visual evidence remains sufficient, as shown in the left region of Figure~\ref{fig:performance_analysis}(c,d). However, once visibility crosses a critical threshold around $0.7$, accuracy drops sharply across settings. This abrupt collapse is consistent with intuition: when key visual evidence is largely unobservable, answering correctly becomes inherently more difficult.

\textbf{Reasoning difficulty drives sustained degradation and poor scaling efficiency.} Despite potential noise in our difficulty metric, the trend is clear: accuracy declines steadily as reasoning difficulty increases, and gains diminish beyond moderate data scale, especially beyond 50K. Because compositional complexity grows rapidly through simple combinations and permutations and can extend far beyond the range shown here, scaling data and model size quickly becomes inefficient. One possible explanation is that current fine-tuning strengthens reasoning chains but struggles with deeper compositional search. Multi-hop reasoning across sparse viewpoints resembles structured graph search that requires exploring alternative relational hypotheses and maintaining cross-view consistency, whereas CoT follows a single linear trajectory that limits exploration and accumulates early errors. These findings suggest that high-difficulty multi-view reasoning may require mechanisms beyond linear CoT, such as structured or tree-based search (\eg. \cite{yao2023treeofthoughts}) to better approximate graph-like relational inference (\eg. \cite{gardenfors2004conceptual}).

\vspace{-1em}

\section{Conclusion}\label{view2space_conclusion}
This work studies multi-view visual reasoning in collaborative settings, where no single observation is sufficient and models must integrate information across discrete viewpoints to form a shared spatial understanding. To support systematic evaluation and scalable training, we develop an automated multi-view data engine with rich geometric and view-consistent annotations, and introduce the VIEW2SPACE benchmark spanning multiple task types, viewpoint configurations, and graded reasoning difficulties. Empirical results show that, while current vision-language models can learn perceptual-level cross-view skills with sufficient data, performance remains limited for high-difficulty multi-hop reasoning and high-occlusion cases, indicating that compositional multi-view reasoning remains an open challenge.


\appendix

\section{Comparison of VIEW2SPACE with existing spatial intelligence benchmarks}

We summarize the key differences between \textsc{VIEW2SPACE} and existing spatial reasoning benchmarks in Table~\ref{tab:diff_benchmark}. In the table, MCQ denotes multiple-choice questioning, where binary indicates questions with only yes/no options, and the remaining cases correspond to standard single-choice questions with multiple candidate answers. ``Count/Free'' refers to counting or free-form answer tasks, which are generally less susceptible to guessing. Detection indicates visual grounding tasks, such as object detection or segmentation. Sparse specifies whether multi-view settings involve sparse viewpoints. FFR stands for free-form reasoning. R-scale denotes the range of reasoning difficulty levels covered by the benchmark. In the metadata column, LM-Gen indicates model-generated annotations, H denotes human annotation, which is typically sparse, with only a small subset of objects in each image annotated and only limited objects provided with orientation labels, and PE denotes physical-engine ground truth, which provides comprehensive annotations for the vast majority of visible objects in the image, including detailed geometric and semantic attributes.

To the best of our knowledge, only a few recent works~\cite{mindcube2025, deng2025internspatial} partially explore multi-view visual reasoning. However, these settings are typically not sparse, and the reasoning depth remains relatively limited. For example, InternSpatial~\cite{deng2025internspatial} primarily focuses on single-view spatial reasoning, with only a limited number of multi-view questions. While MindCube~\cite{mindcube2025} places greater emphasis on multi-view scenarios and introduces explicit representations of inter-view relationships. Its viewpoints are predominantly human-centric and typically exhibit moderate viewpoint variation. The benchmark mainly evaluates perceptual and low-hop spatial reasoning within relatively localized scene regions.
In contrast, \textsc{VIEW2SPACE} emphasizes sparse cross-view inference with greater viewpoint diversity and broader scene coverage, enabling systematic evaluation of compositional multi-hop reasoning under limited visual overlap.

\begin{table*}[ht]
\centering
\caption{Difference between our work and similar benchmarks.MCQ: multiple-choice; Count/Free: counting or free-form answers; Detection: grounding (e.g., detection/segmentation); Sparse: sparse multi-view setups; FFR: free-form reasoning; R-scale: reasoning difficulty range; Metadata: annotation source - LM-Gen (model-generated), H (human), PE (physical engine ground truth).}
\label{tab:diff_benchmark}
{\scriptsize
\resizebox{\linewidth}{!}{\begin{tabular}{cccccccccccc}
\midrule
\rowcolor[gray]{0.95} & Year & Venue & QA Pairs & MCQ & Count/Free & Detection & Multi-view & Sparse & FFR & R-Scale & Metadata\\
\midrule

VSR~\cite{liu2023VSR} & 2023 & TACL & 10K & \textcolor{blue}{Binary} & \textcolor{darkred}{NO} & \textcolor{darkred}{NO} & \textcolor{darkred}{NO} & \textcolor{darkred}{NO} & \textcolor{darkred}{NO} & \textcolor{darkred}{NO} & \textcolor{darkred}{LM-Gen}\\

SpatialRGPT~\cite{cheng2024spatialrgpt} & 2024 & NIPS & 1.4K & \textcolor{darkred}{NO} & \textcolor{darkgreen}{Free} & \textcolor{darkred}{NO} & \textcolor{darkred}{NO} & \textcolor{darkred}{NO}  & \textcolor{darkred}{NO} & \textcolor{darkred}{NO} & \textcolor{darkred}{LM-Gen}\\

VSI-Bench~\cite{yang2025vsi_bench} & 2025 & CVPR & 5.1K & \textcolor{darkgreen}{YES} & \textcolor{darkgreen}{YES} & \textcolor{darkred}{NO} & \textcolor{blue}{Video} & \textcolor{darkred}{NO} & \textcolor{darkgreen}{YES} & \textcolor{darkred}{NO} & \textcolor{darkgreen}{H}\\

OmniSpatial~\cite{jia2026omnispatial} & 2026 & ICLR & 1.5K & \textcolor{darkgreen}{YES} & \textcolor{darkred}{NO} & \textcolor{darkred}{NO} & \textcolor{darkred}{NO} & \textcolor{darkred}{NO} & \textcolor{darkgreen}{YES} & \textcolor{darkred}{NO} & \textcolor{darkred}{LM-Gen}\\

InternSpatial~\cite{deng2025internspatial} & 2026 & ICLR & 12M &\textcolor{darkgreen}{YES} &\textcolor{darkgreen}{YES} &\textcolor{darkgreen}{YES} &\textcolor{blue}{Few} &\textcolor{darkred}{NO}  &\textcolor{darkgreen}{YES} &\textcolor{darkred}{NO} & \textcolor{darkred}{LM-Gen}\\

MindCube~\cite{mindcube2025} & 2026 & ICLR & 20K & \textcolor{darkgreen}{YES} & \textcolor{darkred}{NO} & \textcolor{darkred}{NO} & \textcolor{darkgreen}{YES} & \textcolor{darkred}{NO} & \textcolor{darkgreen}{YES} & \textcolor{darkred}{NO} & \textcolor{darkgreen}{H}\\

VIEW2SPACE(ours) & 2026 & - & 3M & \textcolor{darkgreen}{YES} & \textcolor{darkgreen}{YES} & \textcolor{darkgreen}{YES} & \textcolor{darkgreen}{YES} & \textcolor{darkgreen}{YES} & \textcolor{darkgreen}{YES} & \textcolor{darkgreen}{YES} & \textcolor{darkgreen}{PE}\\

\bottomrule
\end{tabular}}}

\end{table*}

In contrast, real-world spatial reasoning often requires inferring cross-view relationships directly from visual evidence across sparse and heterogeneous viewpoints beyond fixed human perspectives, frequently involving compositional and multi-hop reasoning.

\section{Overview of VIEW2SPACE}
\label{sec:view2space_benchmark_introduction}

\subsection{Thematic Scene Design and Categories}
In this subsection, we provide an overview of the thematic scene design in VIEW2SPACE. The dataset is constructed with an explicit focus on physical plausibility, semantic coherence, and controllable compositional diversity. Each scene theme is manually specified to ensure structurally valid object arrangements and realistic environmental conditions, reducing artifacts commonly introduced by unconstrained procedural generation.

\textsc{VIEW2SPACE} covers over 40 scene themes. Each theme can generate thousands of scene instances with varying scales and layouts, and is defined through a designer-specified configuration file. The configuration encodes the relevant asset categories, admissible spatial relationships and layout patterns, and whether object placement follows deterministic rules or constrained stochastic sampling. Scene-level properties such as floor materials, lighting conditions, and overall scene scale are parametrized within the configuration, enabling systematic variation under controlled constraints.

\begin{figure}[t]
\begin{center}
  \includegraphics[width=0.7\linewidth]{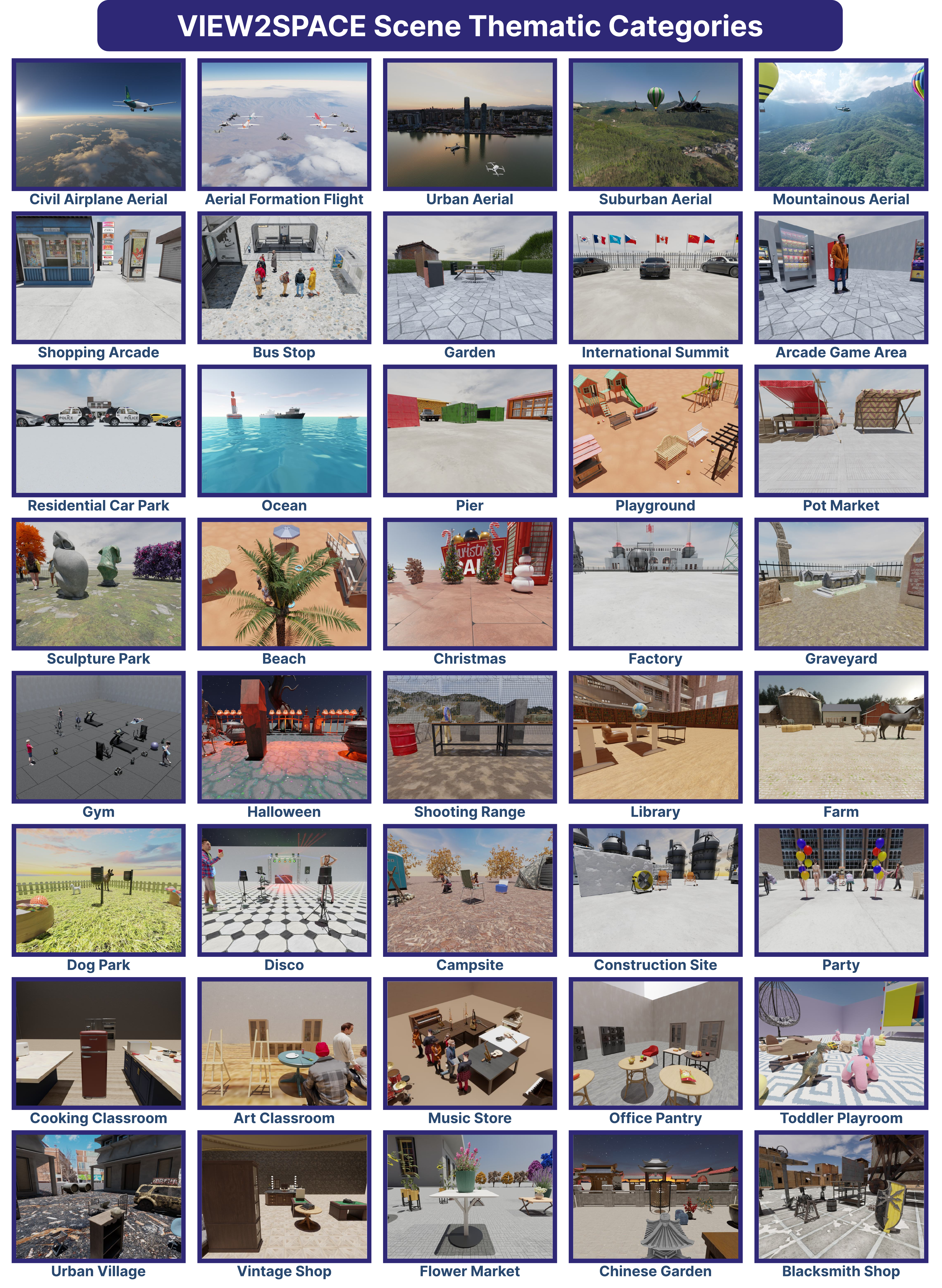} 
\end{center}
\caption{Overview of VIEW2SPACE Scene Thematic Categories}
\label{fig:viewspace_thematic}
\end{figure}

During scene generation, the engine randomly samples scenes within the designer-defined patterns, instantiating varying numbers of objects and scene layouts while respecting the specified constraints. An overview of the VIEW2SPACE scene thematic categories is shown in Figure~\ref{fig:viewspace_thematic}. The resulting scenes are visually realistic, incorporating high-quality 3D assets, including scanned real-world objects, and physically based rendering provided by the underlying physics engine. This design helps reduce the synthetic-to-real gap, enabling models trained on VIEW2SPACE to generalize more effectively to real-world data.

This design substantially reduces manual effort in scene construction. For each scene theme, only a single configuration file needs to be specified, after which the pipeline can automatically generate large numbers of scenes with varying scales, layouts, and object counts. This enables efficient and scalable scene generation without requiring per-scene manual design. \textbf{Under this design, a single scene theme defines a combinatorial generation space, from which an unbounded number of distinct scenes can be sampled by varying object instances, layouts, spatial relations, and scene scale.}

The primary prerequisite for scene generation is the availability of usable 3D assets, after which the pipeline can be immediately deployed. We note that collecting large-scale 3D assets is inherently challenging due to licensing and intellectual property constraints, which limits full automation in practice. As a future direction, we plan to further reduce this bottleneck by incorporating image-to-3D reconstruction techniques (e.g., image-to-point-cloud or image-to-mesh conversion), enabling automatic conversion of 2D images into 3D assets and further improving the scalability and automation of the data generation process.

\begin{figure*}[t]
\begin{center}
  \includegraphics[width=0.8\linewidth]{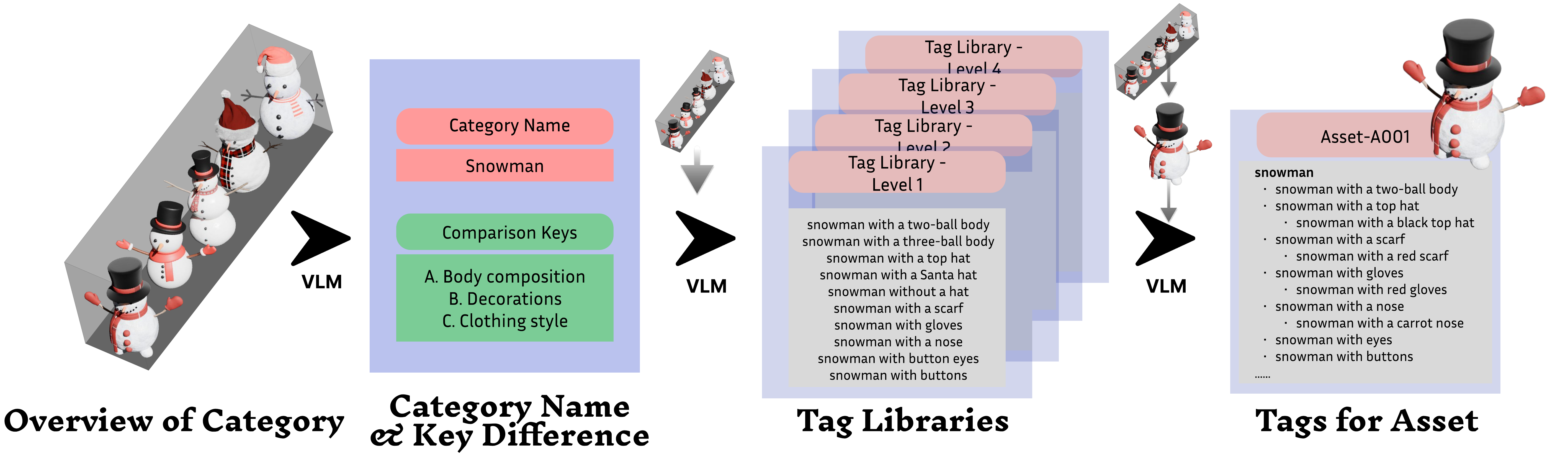} 
\end{center}
\vspace{-1.5em}
\caption{Pipeline of Overview-driven Semantic Differentiation and Tagging}
\label{fig:osg_tag}
\end{figure*}

\subsection{Overview-driven Semantic Differentiation and Tagging (OSD-Tag)}
In this subsection, we introduce our annotation and tagging pipeline for 3D assets, which is designed to capture both shared attributes and fine-grained distinctions among visually and semantically similar assets, as illustrated in Figure~\ref{fig:osg_tag}. Accurate and expressive asset-level annotations are a critical prerequisite for synthetic scene generation, as all downstream processes rely on these tags for scene construction, supervision generation, and question answering.

Our tagging design emphasizes two key principles:
\begin{itemize}
    \item \textbf{Hierarchical richness.} 
    Each asset is annotated with a sufficiently rich set of hierarchical semantic tags, enabling a structured description of its properties at multiple levels of abstraction.
    
    \item \textbf{Cross-asset consistency.} 
    Tags used to describe one asset are carefully verified for transferability to other semantically related assets, ensuring consistent usage across the asset library.
\end{itemize}

\begin{figure*}[!ht]
\begin{center}
  \includegraphics[width=0.7\linewidth]{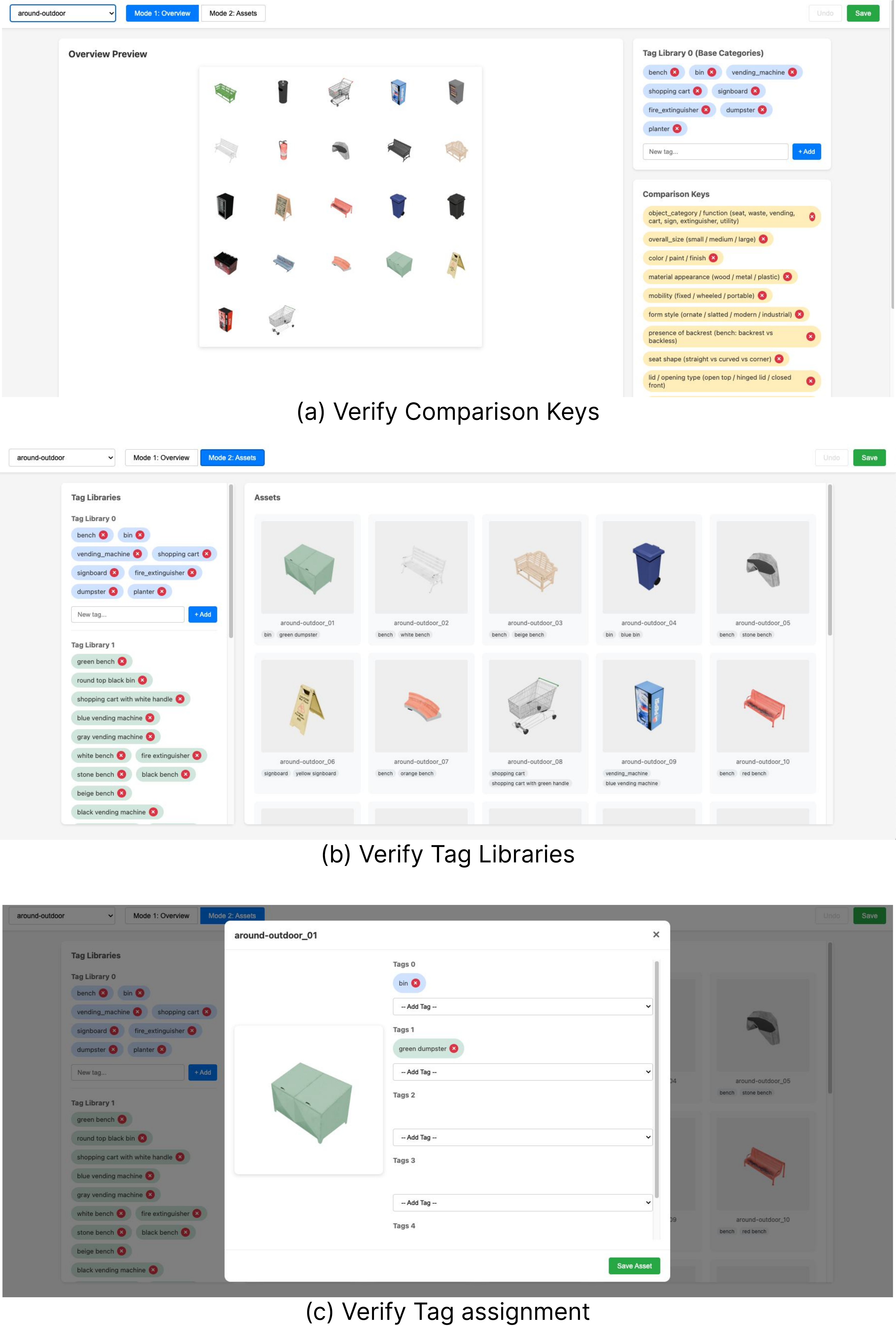} 
\end{center}
\vspace{-1.5em}
\caption{(a) Verify Comparison Keys: check comparison keys across objects within a category; (b) Verify Tag Libraries: check the completeness of the tag library; (c) Verify Tag Assignment: check whether tags are correctly assigned.}
\label{fig:manual_verify}
\end{figure*}

This is crucial for downstream question generation, where asset tags are used to retrieve ground-truth instances. Without consistent and reusable tagging, such retrieval may lead to incorrect or ambiguous ground truth, undermining the reliability of generated supervision.

Motivated by the above considerations, we propose Overview-driven Semantic Differentiation and Tagging (OSD-Tag), a scalable annotation framework designed to satisfy both hierarchical richness and cross-asset consistency. OSD-Tag significantly accelerates the otherwise labor-intensive labeling process while reducing the need for extensive manual verification.

\textbf{OSD-Tag consists of three core modules.}
First, for each asset category, we generate an overview image that visually summarizes representative instances of that category. This single image is used to capture common visual patterns and variations across assets. We then prompt a vision–language model (GPT-5-mini~\cite{singh2025openaigpt5}) with the overview image to produce the category name and identify key semantic dimensions for differentiation among instances. These dimensions include, but are not limited to, structural differences, color schemes, postures, and clothing or appearance attributes, as illustrated in the ``Category Name \& Key Difference'' component of Figure~\ref{fig:osg_tag}.

Second, leveraging the overview image and the comparison keys identified in the first module, we prompt the VLM to analyze both shared components and discriminative attributes among assets within the same category. Based on this analysis, the VLM generates semantic tags at multiple levels of abstraction to describe each asset.
At the first level, tags capture coarse, general properties, such as the presence or absence of major components. The second level provides more concrete and specific descriptions of individual attributes. Higher levels encode increasingly complex combinations of multiple attributes, resulting in a hierarchical tagging structure that progressively refines asset descriptions. Collectively, these multi-level tags form a comprehensive tag library that provides sufficiently rich and expressive annotations across the asset set, as illustrated in the ``Tag Libraries'' component of Figure~\ref{fig:osg_tag}.

Third, to ensure cross-asset consistency, we assign tags to assets by matching each asset against the complete tag library generated in the second module. Rather than allowing tags to be defined independently per asset, every asset is systematically evaluated against the same pool of candidate tags, and all applicable tags are assigned. This design prevents cases where a valid tag applies to multiple assets but is only assigned to a subset of them, which could otherwise lead to incorrect or ambiguous ground truth during downstream question generation, as illustrated in the ``Tags for Asset'' component of Figure~\ref{fig:osg_tag}.

Finally, all generated annotations undergo a rigorous three-step manual verification process (Figure~\ref{fig:manual_verify}), including (a) verification of comparison keys across objects within each category, (b) completeness checks of the tag libraries, and (c) validation of correct tag assignment. This process guarantees consistency and correctness, ensuring high-quality and reliable annotations.

\begin{figure}[t]
\begin{center}
  \includegraphics[width=1.0\linewidth]{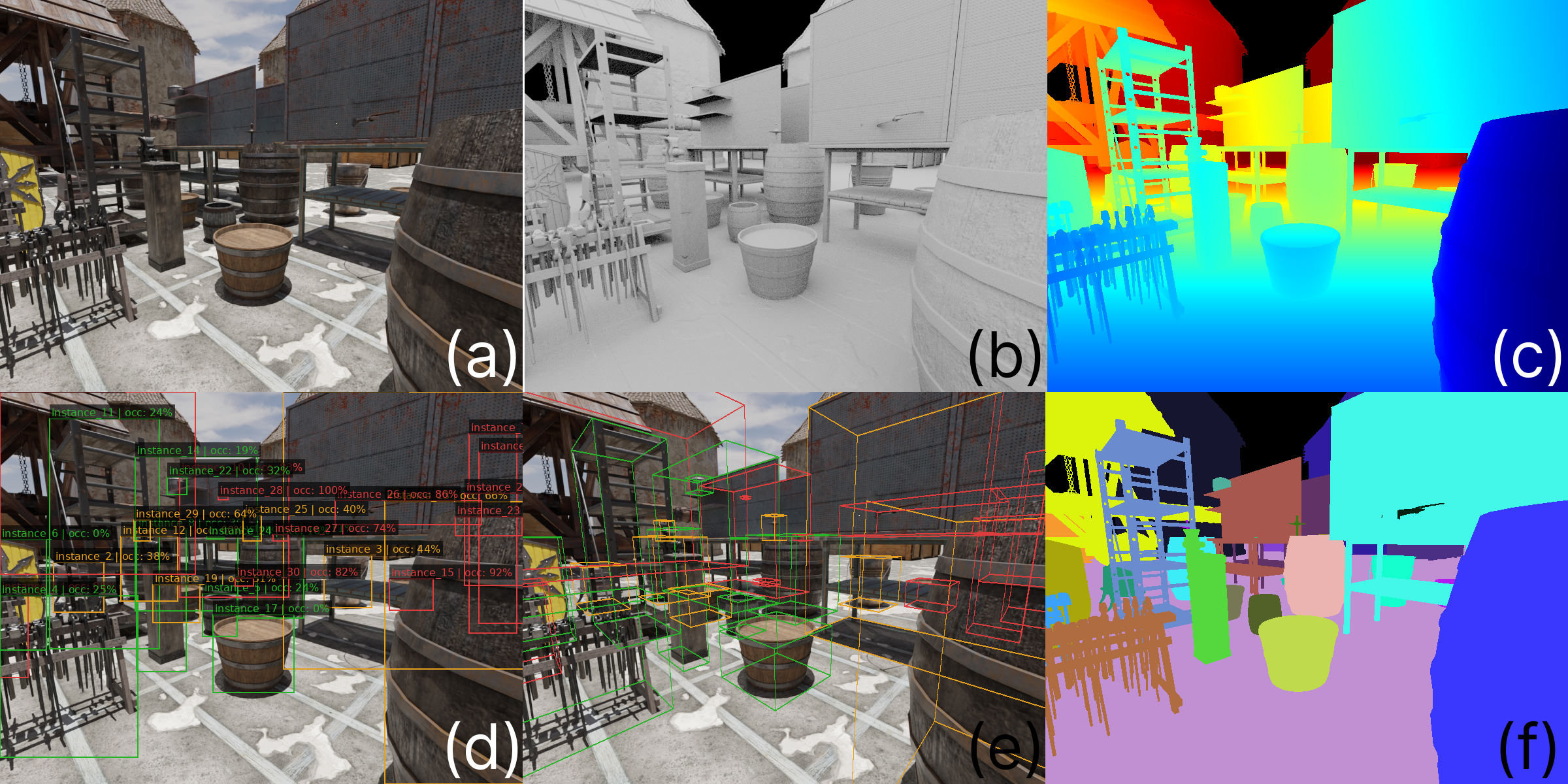} 
\end{center}
\vspace{-1.5em}
\caption{For each RGB image (a), a high-resolution mesh (b) and the corresponding depth map (c) are provided, from which instance occlusion-aware 2D/3D bounding boxes (d/e) and segmentation masks (f) can be readily derived.}
\label{fig:meta_data_vis}
\end{figure}

\subsection{Multi-View Rendering and Metadata Extraction}
To support the diverse requirements of benchmarking, diagnostic analysis, and model training, we design a set of four canonical viewpoint configurations for multi-view rendering, including drone views, bird-eye views, human-like egocentric views, and fixed surveillance camera views. Within each viewpoint category, users can flexibly control camera placement and orientation (\eg, viewing the scene from the center or from peripheral locations, and from different azimuthal directions), as well as camera coverage and optics, such as the proportion of the scene captured and the use of wide-angle versus narrow field-of-view settings. These viewpoints capture scenes from complementary perspectives and introduce systematic variation in viewpoint geometry and visibility.

Using the underlying physics and rendering engines, we automatically compute rich view-dependent metadata, including object bounding boxes, occlusion ratios, and camera-relative visibility statistics as shown in Figure~\ref{fig:meta_data_vis}. We further record precise geometric information for each object, such as 3D pose, orientation, and spatial relationships, which are stored alongside rendered images. This comprehensive metadata enables fine-grained evaluation, controlled diagnostic studies, and the generation of structured supervision for multi-view visual reasoning.

\subsection{Grounded Multi-View Question and Supervision Generation}
We generate question–answer pairs by leveraging the fully controllable scenes and the rich metadata produced by our engine, including object identities, 3D poses, camera parameters, view-dependent bounding boxes, and occlusion ratios. This enables strictly grounded spatial queries whose answers are computed deterministically from geometric ground truth, rather than inferred through probabilistic language model generation or heuristic rules.

We compute object-to-object and object-to-camera spatial relations in both world and camera coordinate frames, forming a deterministic geometric foundation for multi-view reasoning. The process consists of three steps:

\textbf{(1) Geometric relation computation.}
Spatial relations are computed in both object-centric and camera-centric coordinate frames.

\emph{Object-centric relations.}
For objects with a well-defined front orientation (e.g., humans, houses, vehicles), we project the normalized relative displacement vector onto the reference object's local forward and right axes to determine relative direction in the horizontal plane. These projections are discretized into eight canonical directions ( \emph{front},  \emph{back},  \emph{left},  \emph{right}, and their diagonal variants). A projection is treated as zero when its magnitude falls below $\epsilon = 0.1$, corresponding to $|\cos \theta| < 0.1$. Vertical relations such as \emph{on} and \emph{under} are defined by semantic and physics-consistent contact, rather than by vertical displacement alone.

\emph{Camera-centric relations.}
We additionally compute viewer-relative relations (e.g., left, right, front, behind) in the camera coordinate frame.

\textbf{(2) Visibility modelling.}
For each object instance in each view, we record 2D and 3D bounding boxes together with a quantitatively computed occlusion ratio. Visibility is estimated through physics-consistent ray casting in the rendering engine. Specifically, surface points on the object that face the camera are sampled, and rays are emitted toward the camera centre. The occlusion ratio is computed as the fraction of rays that reach the camera without intersecting other geometry, providing a physically grounded measure of visible surface coverage. Additional geometric metadata available from the engine is recorded when needed.

\textbf{(3) Deterministic question instantiation.}
Using the computed geometric relations and visibility metadata, we instantiate rule-based question templates and compute answers deterministically from scene geometry.

These settings further enable structured supervision, including instance-level grounding with identities and tags, explicit inter-view relationship, and multi-step reasoning traces derived from deterministic geometric ground truth.

\section{VIEW2SPACE-V1 Answer Distribution and Statistical Analysis}

\begin{figure}[ht]
\begin{center}
  \includegraphics[width=0.98\linewidth]{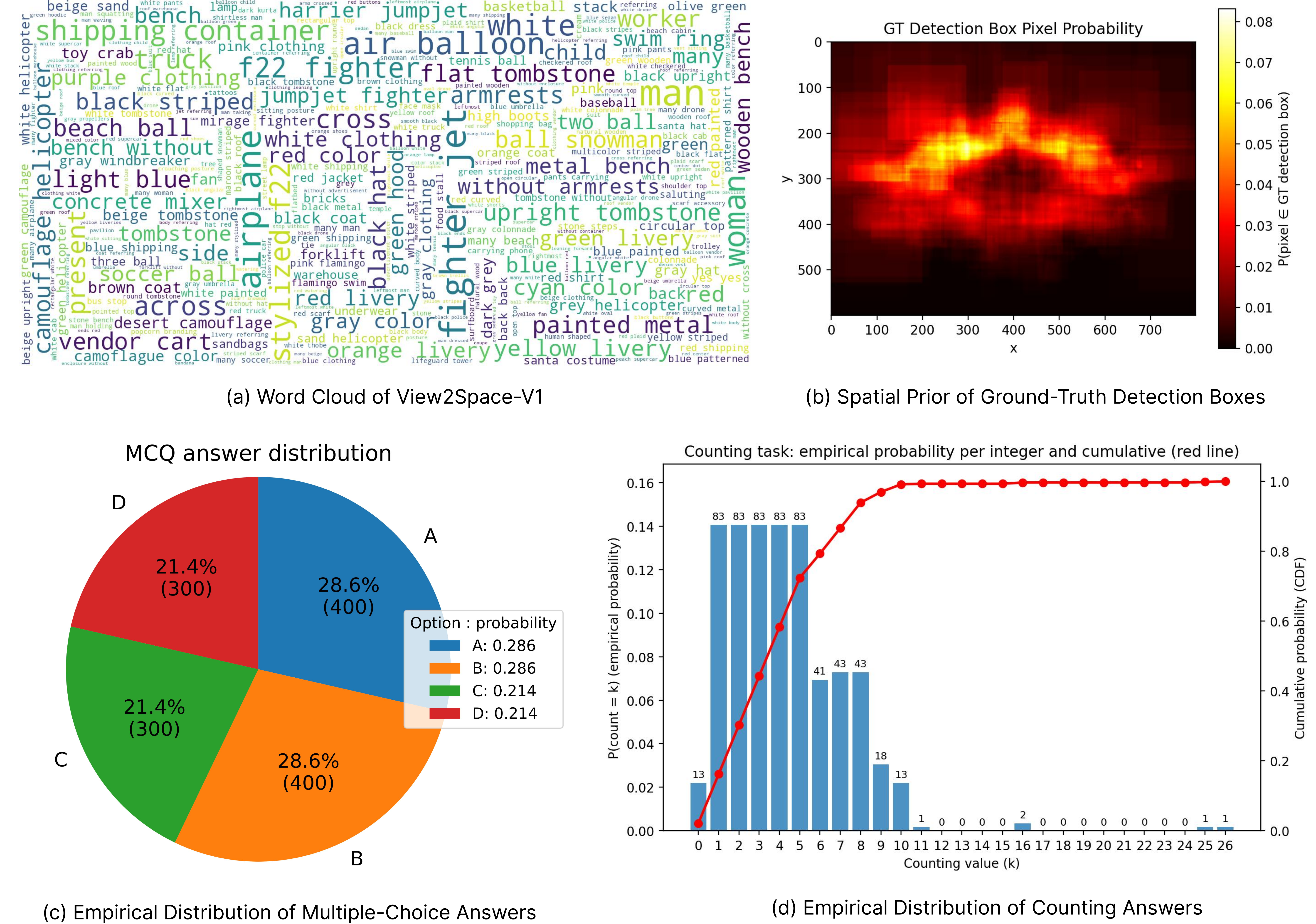} 
\end{center}
\vspace{-1.5em}
\caption{Answer Distribution and Statistical Analysis of View2Space-V1}
\label{fig:data_stats_vis}
\end{figure}

Figure~\ref{fig:data_stats_vis} presents a comprehensive analysis of the answer distributions in View2Space-V1 across multiple task types. Overall, the distributions exhibit natural and well-balanced characteristics, indicating that the dataset does not suffer from obvious answer or spatial biases.

The word cloud in Figure~\ref{fig:data_stats_vis} (a) highlights a rich and diverse vocabulary spanning objects, attributes, and scene-level concepts, reflecting the broad visual–linguistic coverage of View2Space-V1. This diversity suggests that the dataset captures a wide range of semantic concepts rather than being dominated by a small set of frequent terms. We note that the vocabulary will continue to expand as we iteratively extend and refine the dataset in future releases.

Figure~\ref{fig:data_stats_vis} (b) visualizes the pixel-wise probability of ground-truth detection bounding boxes, revealing a relatively uniform spatial distribution across the image plane. Unlike datasets where annotations are heavily concentrated in a limited region, View2Space-V1 exhibits a balanced spatial prior, reducing location-specific biases and encouraging models to reason over the entire visual field.

For multiple-choice questions, Figure~\ref{fig:data_stats_vis} (d) shows that answer options are approximately evenly distributed, as the choices are randomly shuffled during data construction. This design prevents systematic biases toward specific options and ensures that models cannot exploit answer-position shortcuts.

Finally, Figure~\ref{fig:data_stats_vis} (d) reports the empirical distribution of counting task answers. The counts are spread across a range of values rather than collapsing to a single dominant number, indicating that the counting questions exhibit sufficient variability and complexity.

Looking forward, we plan to further enhance View2Space by incorporating more challenging cases, including scenarios that may induce visual or reasoning-based hallucinations. These extensions will be released in subsequent versions to support more robust evaluation of multimodal models.

\section{Reasoning Difficulty Definition}
We explicitly quantify reasoning difficulty through a graph-based formulation inspired by prior structured reasoning frameworks~\cite{kim2025biohopr, koner2021graphhopper, yang2020graph}, while adapting the formulation to multi-view spatial inference. For each question, we decompose the required reasoning process into per-view sub-graphs, where key objects are represented as nodes and spatial relations as edges. Solving the question corresponds to traversing the sub-graph starting from a reference node toward the target node(s).

We define difficulty as the minimum number of reasoning hops required to reach the target node(s). A single perceptual grounding step (i.e., mapping visual input to an object node) is counted as one hop. Multi-step relational inference increases the hop count accordingly.

To account for combinatorial scene complexity, we further incorporate a structural factor proportional to $\log_2(N)$, where $N$ is the number of key objects involved in the reasoning chain. The final difficulty score thus reflects both relational depth (multi-hop reasoning) and object-level
combinatorial complexity.

We note that this hop-based formulation provides a structured and interpretable approximation of reasoning complexity, but it does not fully capture all sources of cognitive difficulty. For example, a single-hop camera rotation reasoning and a single-hop relational query may both be counted as one hop, although their practical difficulty can differ. Reducing such measurement noise and developing more fine-grained difficulty metrics remains an important direction for future work, which may further improve difficulty-aware data analysis.

\begin{table}[ht]

\caption{Thinking vs. Non-Thinking Comparison on \textsc{VIEW2SPACE-v1}. The light blue region labelled ``Direct'' corresponds to inference without chain-of-thought reasoning, where the model produces the final answer directly.}
\label{tab:cot_vs_direct}
\centering
\scalebox{0.8}{
\begin{tabular}{lc|cc|cc}
\toprule[0.3mm]
\rowcolor[gray]{0.9} Method & \multicolumn{1}{c|}{Multiple Choice Answering} & \multicolumn{2}{c|}{Visual Counting } & \multicolumn{2}{c}{Visual Grounding} \\
\rowcolor[gray]{0.9}& ACC (\%) & MAE & ACC (\%) & mIoU (\%) & F1-score (\%)\\

\rowcolor[gray]{0.97} Baseline & & & & &\\
Random (chance)         & 28.59 & 11.37 & 3.39 & 0.22 & 0.0\\
Random (frequency)      & 28.59 & 2.46 & 13.05 & 2.01 & 0.27 \\
\hline
\rowcolor[gray]{0.97} Open-source MLLMs  & & & & &\\


Mantis-8B (SigLip)~\cite{jiangmantis} & 30.07 & 2.94 & 17.60 & 1.24 & 0.23 \\

\rowcolor{blue!5} Mantis-8B (SigLip)-Direct~\cite{jiangmantis} & 30.21 & 2.91 & 15.91 & 0.02 & 0.0 \\

DeepSeek-VL2-Small~\cite{lu2024deepseekvl} & 29.43 & 2.20 & 24.37 & 2.86 & 0.19 \\

\rowcolor{blue!5} DeepSeek-VL2-Small-Direct~\cite{lu2024deepseekvl} & 32.00 & 2.50 & 21.36 & 4.68 & 0.24 \\

Gemma-3-12B-it~\cite{team2025gemma3} & 31.14 & 2.08 & 24.53 & 2.49 & 0.24 \\

\rowcolor{blue!5}Gemma-3-12B-it-Direct~\cite{team2025gemma3} & 32.50 & 2.24 & 25.81 & 2.41 & 0.27 \\

Idefics2-8B~\cite{laurenccon2024idefics2} & 26.64 & 3.92 & 11.34 & 0.39 & 0.0 \\

\rowcolor{blue!5}Idefics2-8B-Direct~\cite{laurenccon2024idefics2} & 25.57 & 3.85 & 12.35 & 0.28 & 0.0 \\


Qwen3-VL-2B~\cite{Qwen3VL}    & 31.43 & 2.63 & 16.34 & 11.20 & 5.07 \\

\rowcolor{blue!5}Qwen3-VL-2B-Direct~\cite{Qwen3VL}    & 31.71 & 1.88 & 22.34 & 14.15 & 4.87 \\

Qwen3-VL-4B~\cite{Qwen3VL}    & 35.19 & 2.26 & 21.15 & 16.53 & 18.72\\

\rowcolor{blue!5}Qwen3-VL-4B-Direct~\cite{Qwen3VL}    & 32.79 & 2.10 & 23.69 & 19.47 & 11.95 \\

Qwen3-VL-8B~\cite{Qwen3VL}    & 37.41 & 2.10 & 24.92 & 10.05 & 7.37 \\

\rowcolor{blue!5}Qwen3-VL-8B-Direct~\cite{Qwen3VL}    & 31.79 & 2.51 & 20.98 & 14.23 & 9.69 \\
\hline
\rowcolor[gray]{0.97} Open-source Spatial Models  & & & & &\\
Molmo2-4B~\cite{clark2026molmo2}   & 31.36 & 2.21 & 22.17 & 1.17 & 0.13 \\

\rowcolor{blue!5}Molmo2-4B-Direct~\cite{clark2026molmo2}   & 30.64 & 2.54 & 22.67 & 1.11 & 0.19 \\

Molmo2-8B~\cite{clark2026molmo2}   & 34.57 & 5.93 & 21.66 & 1.51 & 0.0  \\

\rowcolor{blue!5}Molmo2-8B-Direct~\cite{clark2026molmo2}   & 31.00 & 2.65 & 20.14 & 0.97 & 0.0 \\

RoBobrain2.0~\cite{team2025robobrain2}& 30.79 & 2.22 & 21.49 & 0.0 & 0.0\\

\rowcolor{blue!5}RoBobrain2.0-Direct~\cite{team2025robobrain2}& 29.92 & 2.21 & 21.15 & 0.0 & 0.0\\

Spatial-MLLM~\cite{wu2025spatialmllm}& 27.93 & 2.10 & 9.98 & 0.0 & 0.0\\
\rowcolor{blue!5}Spatial-MLLM-Direct~\cite{wu2025spatialmllm}& 28.00 & 2.08 & 10.83 & 0.0 & 0.0\\

Openview~\cite{chen2025openview} & 30.72 & 2.57 & 19.80 & 1.08 & 0.30\\
\rowcolor{blue!5}Openview-Direct~\cite{chen2025openview} & 30.72 & 2.57 & 18.61 & 2.93 & 0.35\\

MINDCUBE-Aug~\cite{mindcube2025}& 29.00 & 4.52 & 18.10 & 0.0 & 0.0 \\

\rowcolor{blue!5}MINDCUBE-Aug-Direct~\cite{mindcube2025}& 29.00 & 2.16 & 19.97 & 0.06 & 0.0 \\

MINDCUBE-Plain~\cite{mindcube2025}& 30.21 & 4.77 & 19.12 & 0.12 & 0.0 \\

\rowcolor{blue!5}MINDCUBE-Plain-Direct~\cite{mindcube2025}& 29.71 & 2.17 & 21.23 & 0.06 & 0.0 \\
\hline

\bottomrule[0.3mm]
\end{tabular}}
\end{table}

\section{Thinking vs. Non-Thinking Inference on VIEW2SPACE.}
This section provides detailed results corresponding to Section 4. In the main evaluation, models were prompted without restricting reasoning style, thereby allowing them to invoke explicit thinking when supported. For controlled comparison, we construct a direct-answer variant that explicitly instructs models to skip intermediate reasoning. In Table~\ref{tab:cot_vs_direct}, entries marked with “-Direct” denote this direct-answer prompting.

Consistent with the main findings, explicit thinking does not consistently improve performance. While the impact is relatively modest on simpler spatial tasks, several open-source VLMs, particularly the Qwen3VL series~\cite{Qwen3VL}, exhibit more pronounced differences between thinking and direct-answer inference. In certain tasks, enforced CoT leads to lower accuracy than direct-answer prompting.

Manual inspection further clarifies this behaviour. In sparse multi-view settings, direct answers sometimes succeed by exploiting shortcut cues. When thinking is enforced, reasoning traces frequently become lengthy and repetitive, with limited integration of cross-view evidence, and often conclude with incorrect or undecided outputs. These observations suggest that without sufficient multi-view training, explicit reasoning may propagate weak spatial representations rather than refine them, reducing the potential advantage of shortcut-based inference.

\begin{table}

\caption{Human Evaluation on VIEW2SPACE.}
\label{tab:human_eva}
\centering
\scalebox{0.8}{
\begin{tabular}{lc|c|c}
\toprule[0.3mm]
\rowcolor[gray]{0.9} Answerer & \multicolumn{1}{c|}{Multiple Choice Answering} & \multicolumn{1}{c|}{Visual Counting } & \multicolumn{1}{c}{Visual Grounding} \\
\rowcolor[gray]{0.9}& ACC (\%) & ACC (\%) & mIoU (\%)\\

\rowcolor[gray]{0.97} Baseline  & & & \\
Random (chance)         & 28.59  & 3.39 & 0.22 \\
Random (frequency)      & 28.59  & 13.05 & 2.01 \\

\hline
\rowcolor[gray]{0.97} Vision Language Models  & & & \\
Qwen3-VL-4B~\cite{Qwen3VL}    & 35.19  & 21.15 & 16.53 \\
GPT-5~\cite{singh2025openaigpt5}  & 59.86  & 38.10 & 8.18 \\

\hline
\rowcolor{TrainedRow} Fine-tuned Model (Qwen3VL-4B) & & &  \\

Ours (Grounded CoT) & 64.93 & 54.99 & 69.34 \\

\rowcolor{blue!5} Human & & & \\
Human-avg    & 89.88  & 86.24 & 81.62 \\
Human-min    & 85.00  & 81.05 & 77.53 \\
Human-max    & 93.57  & 91.37 & 85.67 \\

\bottomrule[0.3mm]
\end{tabular}}
\end{table}

\section{Human Evaluation}
We conduct a human evaluation study on \textsc{VIEW2SPACE-v1}, with results summarized in Table~\ref{tab:human_eva}. The evaluation spans all task categories and is performed by five annotators at undergraduate or master’s level. Each annotator independently answers the questions without collaboration.

Human performance consistently surpasses current vision-language models. On MCQ tasks, the average performance gap between GPT-5 and human participants is approximately 30\%. The disparity becomes substantially larger for detection tasks. While human annotators achieve an average mIoU of 81.62\%, the best-performing detection model, Qwen3-VL-4B~\cite{Qwen3VL}, achieves 16.53\%, revealing a significant performance gap in fine-grained grounding.

Interestingly, the gap between humans and trained models is smaller than that observed for general-purpose models on detection tasks. Manual inspection of responses suggests that localizing small objects often requires zooming and careful visual search, and the annotated bounding regions are typically small. In practice, some annotators provide approximate rather than strictly precise bounding boxes, which may partially narrow the measured gap. Nevertheless, a margin of around 10\% remains between humans and the strongest trained models, indicating that substantial room for improvement persists in grounding-intensive multi-view reasoning.

\begin{figure*}[ht]
\begin{center}
  \includegraphics[width=0.8\linewidth]{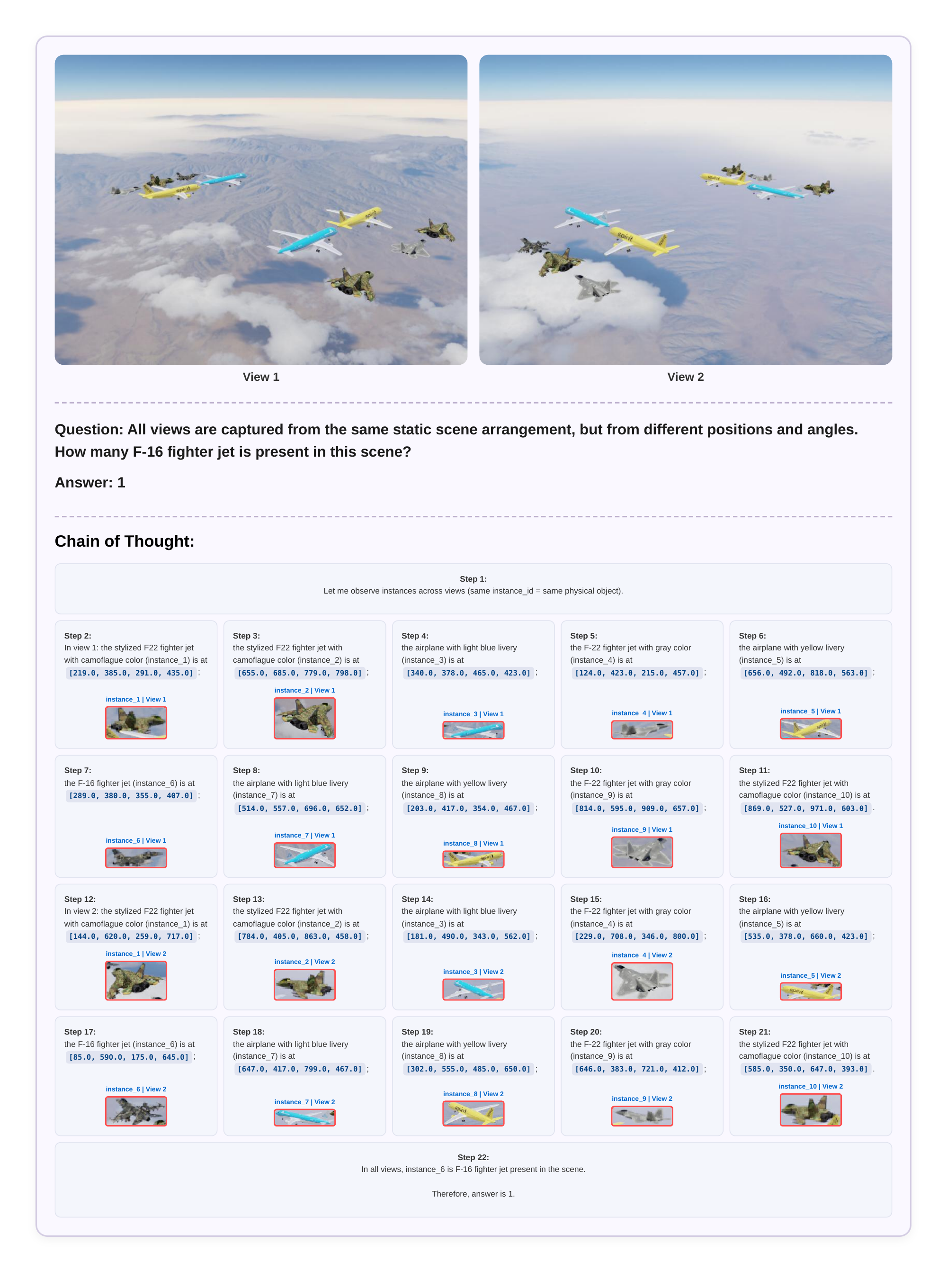} 
\end{center}
\vspace{-3em}
\caption{Sample of training data - 1}
\end{figure*}

\section{A Scalable Training Split for VIEW2SPACE}
We construct a scalable training dataset comprising 300K samples, with scenes disjoint from the evaluation split and support for expansion to millions of question–answer pairs. Each instance is derived from engine-level geometric and physical ground truth, including bounding boxes, occlusion ratios, and precise cross-view spatial relations.

Through a symbolic generation pipeline, we produce deterministic and visually grounded supervision signals with verifiable intermediate reasoning steps. The supervision can be configured at multiple levels of granularity. Specifically, for each training sample, we can output: (1) the final answer only; (2) textual chain-of-thought reasoning; (3) explicit cross-view relational descriptions; and (4) object-level visual evidence, including 2D bounding boxes for each referenced object in the corresponding views. All questions in the training set are generated programmatically and can be deterministically reconstructed.

If desired, visual evidence can be further converted into alternative representations such as object centre coordinates or segmentation masks. In this work, we adopt 2D bounding boxes as the grounding supervision, which already yields strong empirical performance. We provide qualitative examples below to illustrate the structure of the generated supervision.

\section{Supervised Fine-Tuning Setting on VIEW2SPACE}

\subsection{Implementation Details}
We fine-tune three supervised variants on the 300k \textsc{VIEW2SPACE} training set under the same training protocol: (1) instruction-tuning, (2) supervised chain-of-thought (CoT) training, and (3) our proposed Grounded Chain-of-Thought with Visual Evidence (Grounded-CoT). All variants are trained using the HuggingFace TRL library with full-parameter optimization; no parameter-efficient adapters such as LoRA are applied.

Training is conducted using the HuggingFace TRL library with full-parameter optimization; no parameter-efficient adaptation methods such as LoRA are used. Models are trained for 2 epochs with a per-device batch size of $2$ and gradient accumulation over 32 steps. The learning rate is set to $3\times10^{-5}$ with a warmup ratio of 0.05.
Unless otherwise specified, all other hyper-parameters follow the default settings of the corresponding model implementations. We will publicly release the training code, configuration files, and model checkpoints to facilitate reproducibility and support future research on multi-view spatial reasoning.

\begin{figure}[ht]
\begin{center}
  \includegraphics[width=0.98\linewidth]{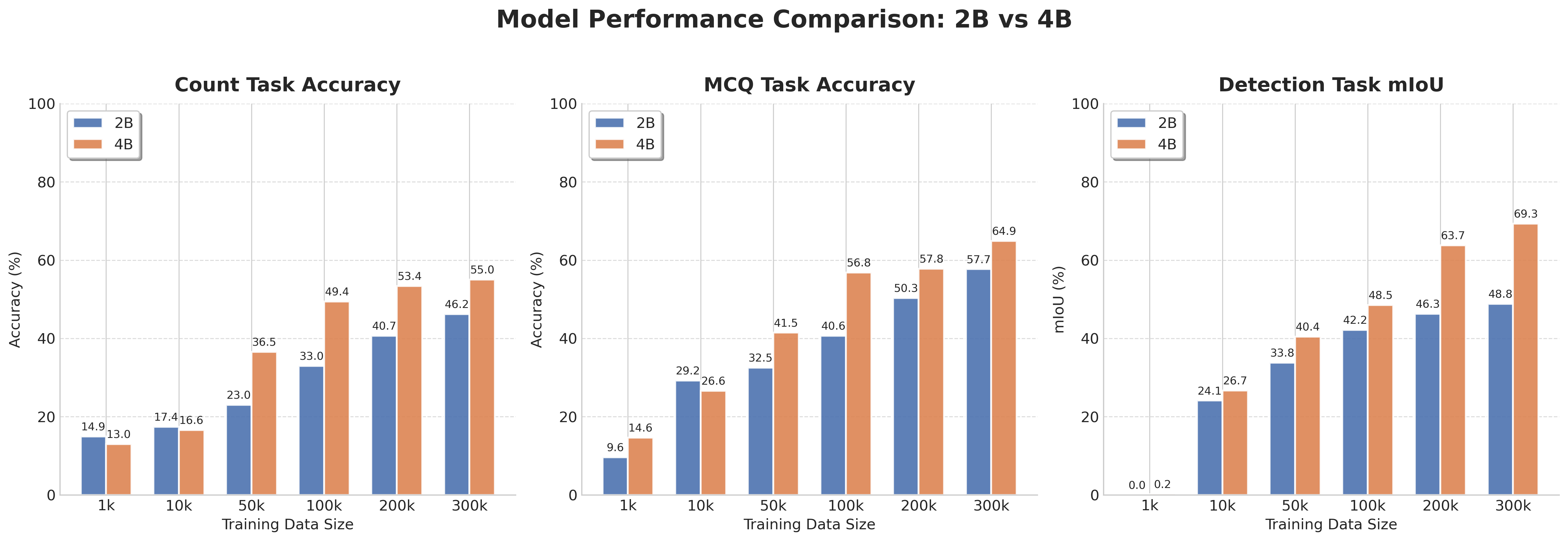} 
\end{center}
\vspace{-1.5em}
\caption{Scaling behavior of Qwen3-VL-2B and Qwen3-VL-4B under \emph{Grounded Chain-of-Thought with Visual Evidence} training on \textsc{VIEW2SPACE}. Performance is reported across increasing training data sizes (1k–300k) for count (accuracy), MCQ (accuracy), and detection (mIoU) tasks. While both models benefit from additional data, MCQ performance exhibits diminishing returns beyond 100k samples—particularly for the 4B model—whereas detection tasks show a clear data threshold effect, with near-zero performance at very small scales and substantial gains once sufficient grounded supervision is provided.}
\label{fig:2b4bcomparision}
\end{figure}

\begin{figure*}[ht]
\begin{center}
  \includegraphics[width=0.95\linewidth]{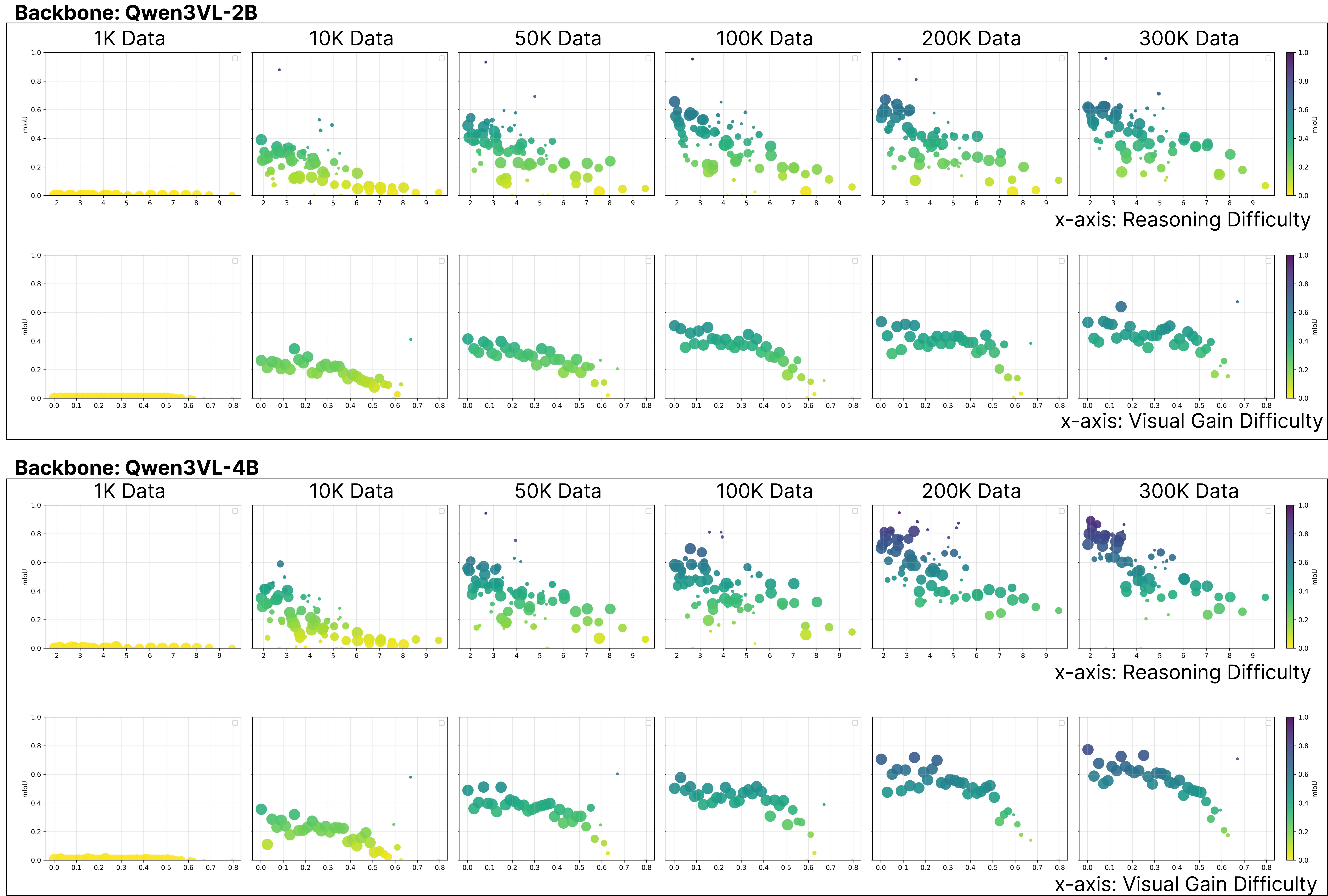} 
\end{center}

\caption[Scaling behaviour on the detection task.]{Scaling behaviour across dataset size and difficulty. Each subplot reports mIoU. Empirical data points over dataset sizes (2B and 4B) are shown here.}
\label{fig:performance_analysis2}

\end{figure*}

\section{Scaling Behaviour under Grounded Chain-of-Thought Training using our engine}

To better understand how training data scale affects multi-view reasoning, we conduct a controlled study using Qwen3-VL-2B~\cite{Qwen3VL} and Qwen3-VL-4B~\cite{Qwen3VL}. For each model, we fine-tune under our \emph{Grounded Chain-of-Thought with Visual Evidence} framework using progressively larger subsets of engine-generated data, ranging from 1k to 300k samples (1k, 10k, 50k, 100k, 200k, 300k). All models are evaluated on the full \textsc{VIEW2SPACE-v1} benchmark as illustrated in Figure ~\ref{fig:2b4bcomparision}.

\subsection{Scaling Behavior Across Visibility Difficulty}

Scaling data and model size improves performance across a broad range of visibility difficulty. Accuracy declines gradually as visibility decreases, indicating that current models can learn effective geometric and perceptual representations when sufficient visual evidence is present. However, when visibility exceeds approximately 0.7, performance drops sharply across all settings. This behavior suggests an observability boundary: once key visual evidence becomes largely inaccessible, performance degrades regardless of additional data or model capacity.

\subsection{Scaling Behavior Across Reasoning Difficulty}

In contrast, reasoning difficulty exhibits sustained degradation and diminishing scaling returns. Accuracy decreases steadily as hop complexity increases, and gains become marginal beyond moderate data scale, particularly after 50K samples. While data scaling raises overall performance levels, it does not fundamentally alter the slope of degradation under increasing reasoning complexity.

This pattern likely reflects structural limitations of linear chain-of-thought supervision. Multi-hop reasoning across sparse viewpoints resembles structured graph search that requires maintaining cross-view consistency and exploring alternative relational hypotheses. Standard CoT fine-tuning follows a single linear trajectory, which may limit exploration depth and propagate early inference errors. As compositional complexity grows combinatorially, scaling data and model size alone becomes increasingly inefficient.

Overall, these results indicate two distinct regimes: visibility difficulty corresponds to a learnable perceptual regime bounded by observability constraints, whereas reasoning difficulty exposes structural limitations in compositional inference. Addressing high-difficulty multi-view reasoning may therefore require mechanisms beyond linear CoT, such as structured or tree-based search strategies~\cite{yao2023treeofthoughts} that better approximate graph-like relational reasoning~\cite{gardenfors2004conceptual}.

\section{Ethics Statement}
The dataset used in this work does not contain personally identifiable information and does not involve human participants, sensitive personal records, or protected data. All materials are sourced from publicly accessible resources or obtained under appropriate usage agreements, and are utilized solely for academic research. Data management procedures are implemented to preserve confidentiality and maintain data integrity throughout the study. Accordingly, the research does not present identifiable ethical risks to individuals or broader society.

\section{VIEW2SPACE Examples}
We present representative qualitative examples from \textsc{VIEW2SPACE} to illustrate the range of question types and reasoning patterns covered in the benchmark. These examples highlight the diversity of spatial configurations, viewpoint variations, and grounding requirements involved in multi-view reasoning.

\begin{figure*}[ht]
\begin{center}
  \includegraphics[width=1.0\linewidth]{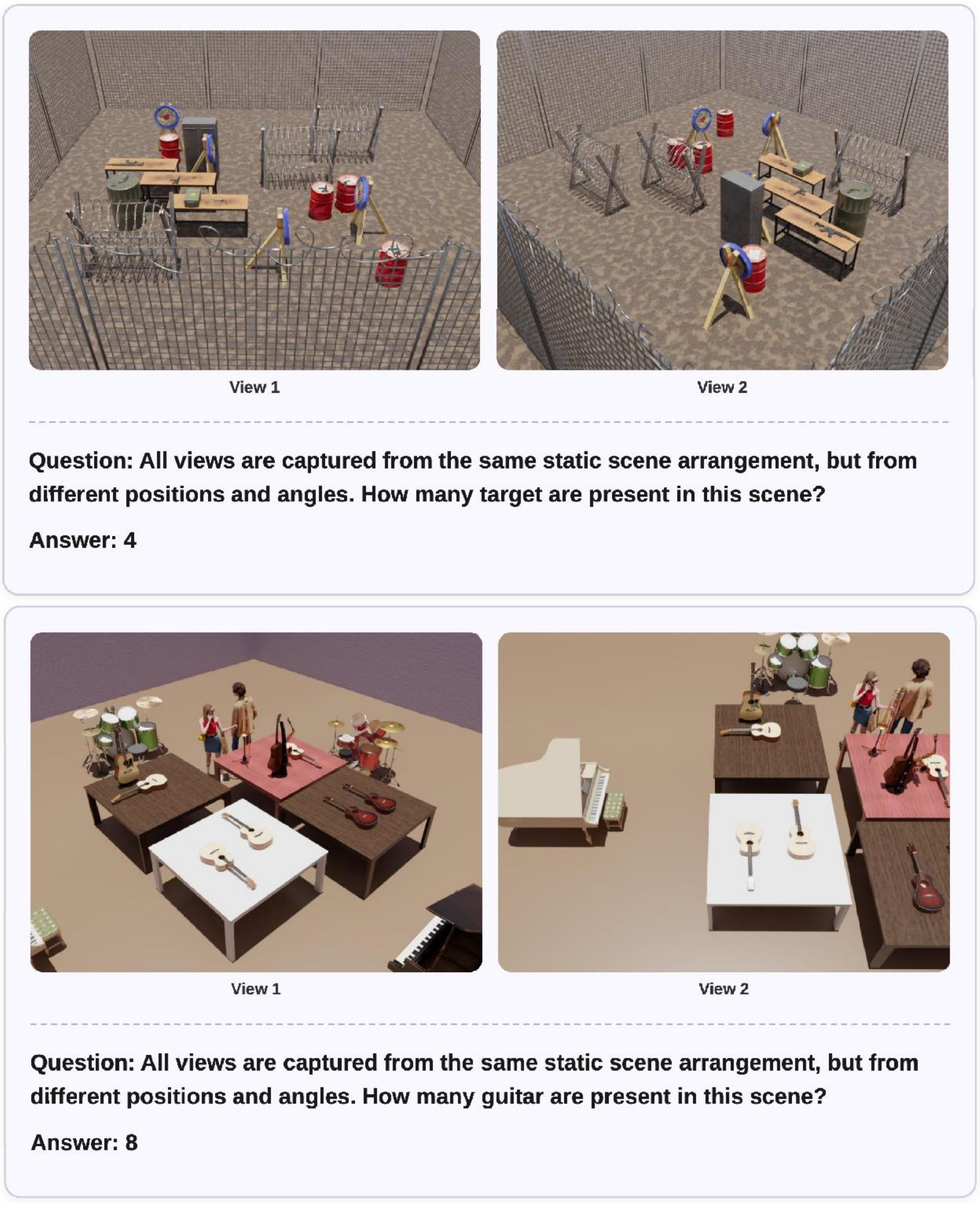} 
\end{center}
\vspace{-3em}
\end{figure*}

\begin{figure*}[ht]
\begin{center}
  \includegraphics[width=1.0\linewidth]{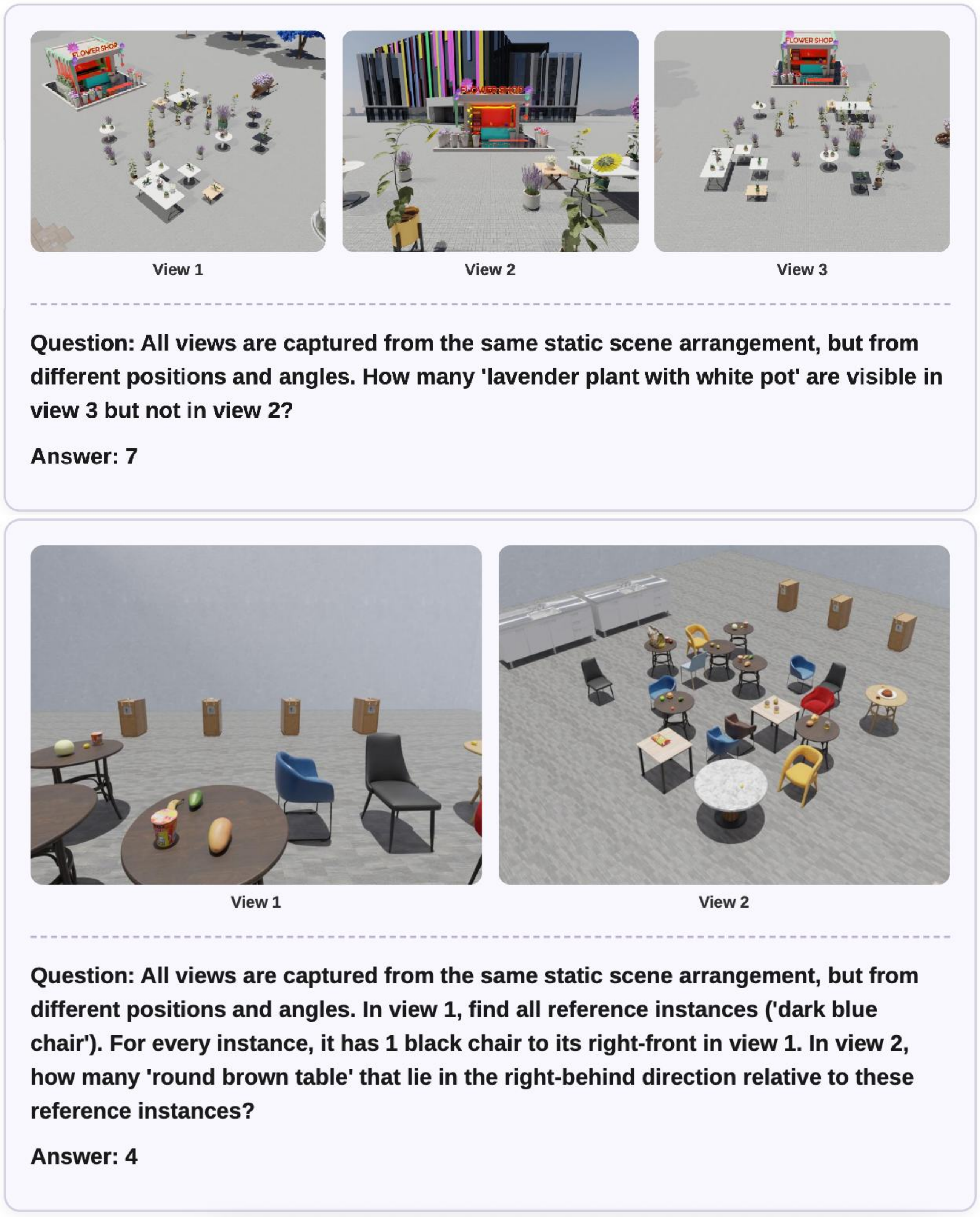} 
\end{center}
\vspace{-2em}
\caption{Samples of counting task}
\end{figure*}

\begin{figure*}[ht]
\begin{center}
  \includegraphics[width=1.0\linewidth]{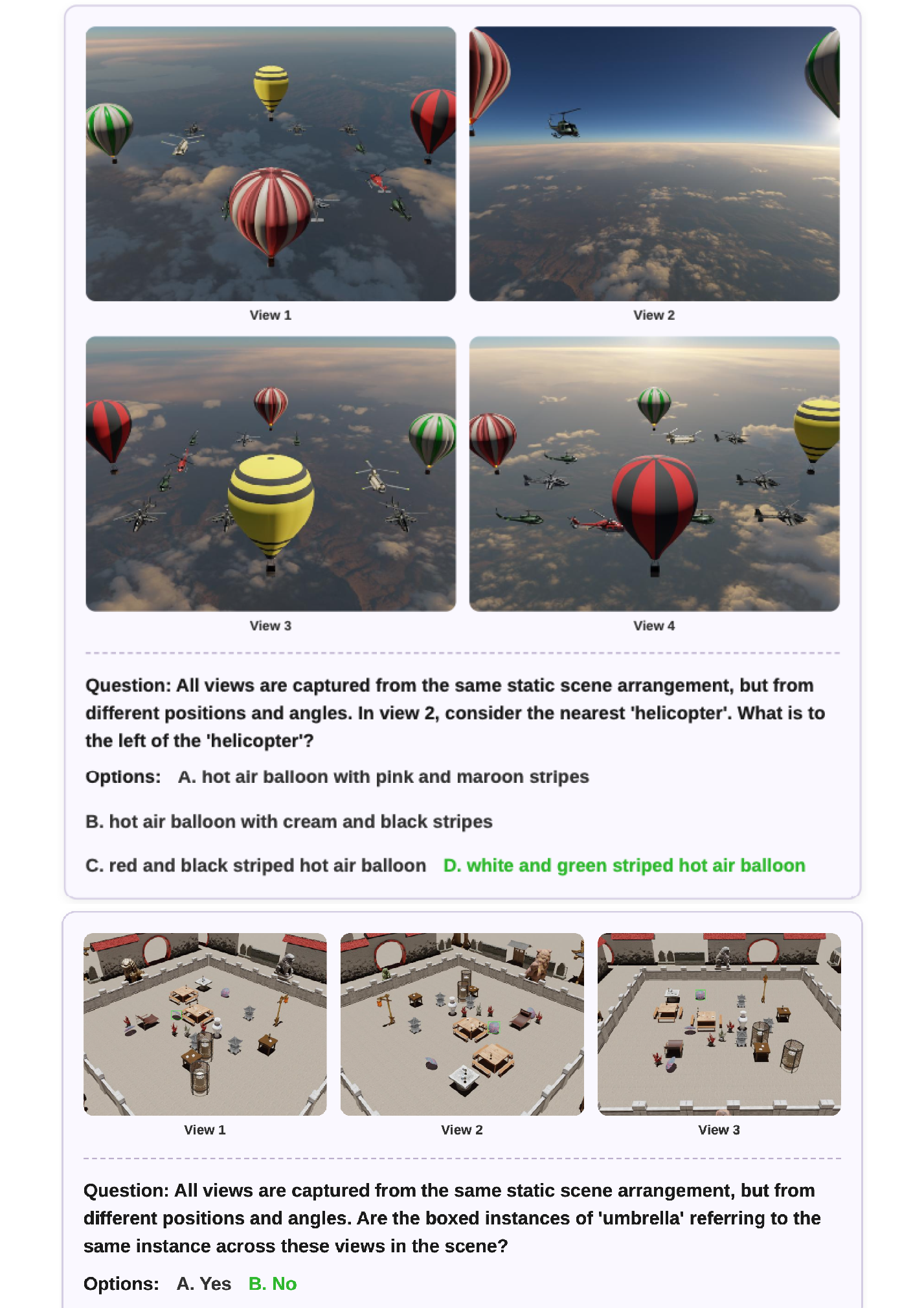} 
\end{center}
\end{figure*}

\begin{figure*}[ht]
\begin{center}
  \includegraphics[width=1.0\linewidth]{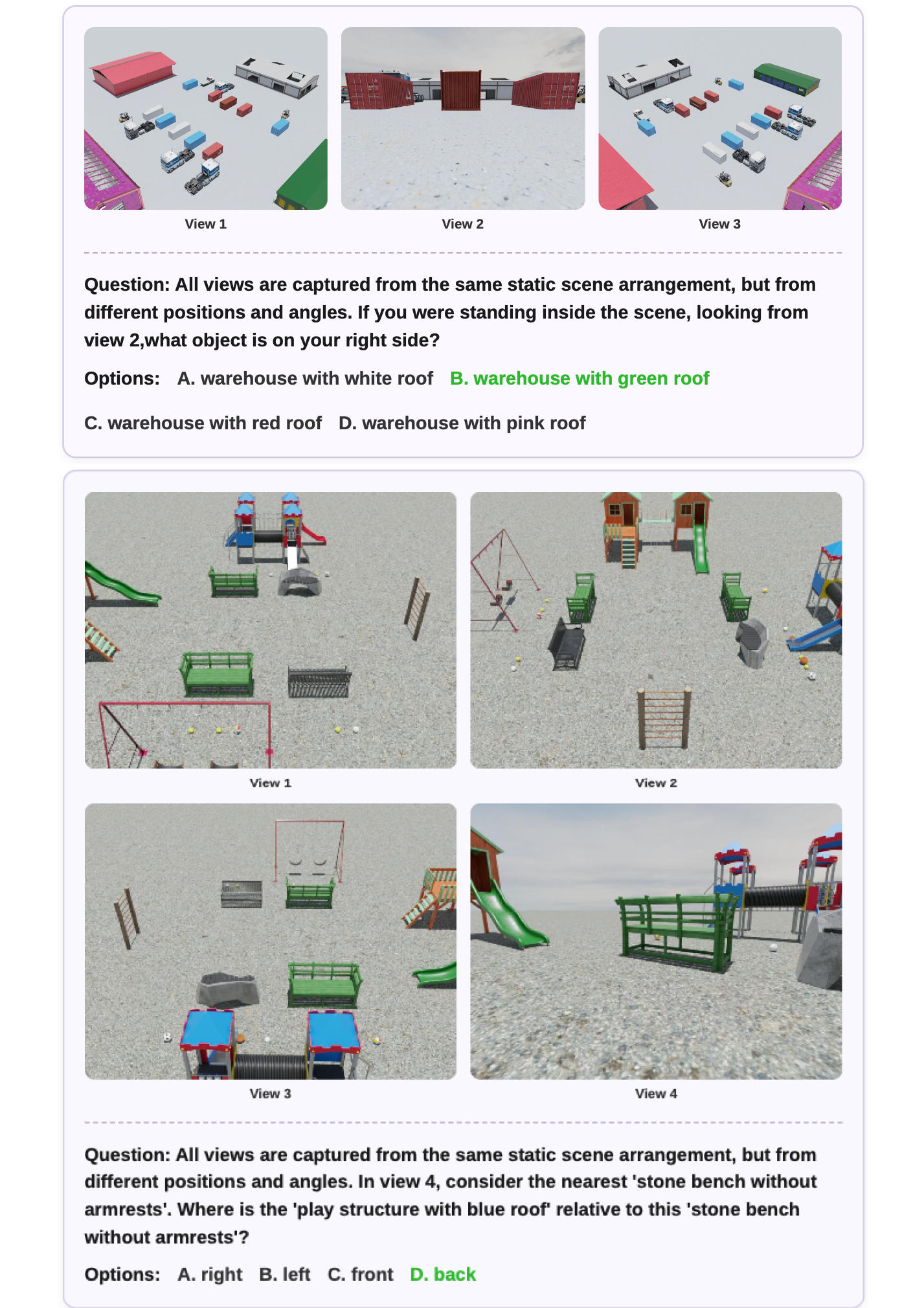} 
\end{center}
\label{fig:sample_of_mcq}
\caption{Samples of multiple choice questioning tasks}
\end{figure*}

\begin{figure*}[ht]
\begin{center}
  \includegraphics[width=1\linewidth]{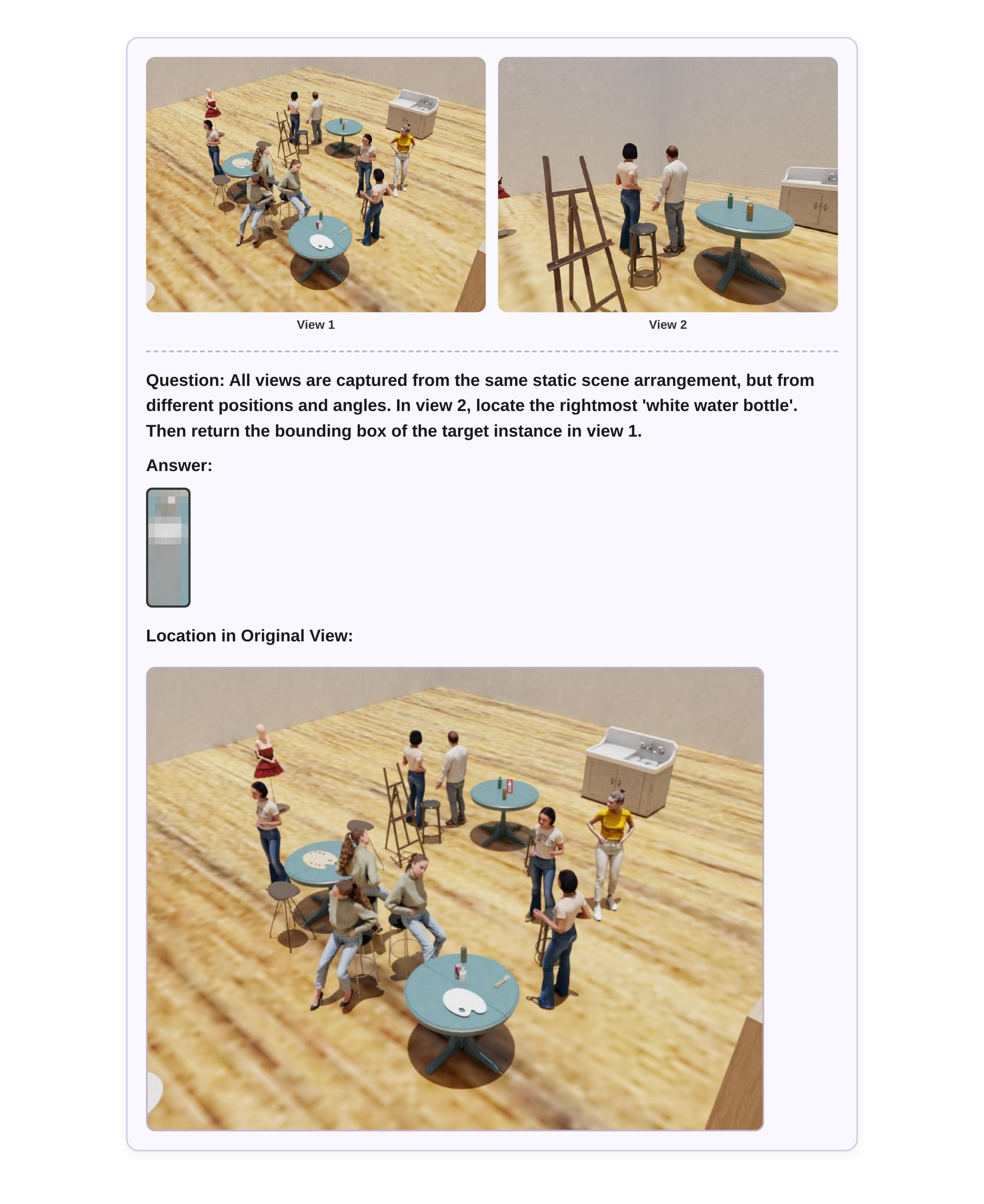} 
\end{center}
\vspace{-3em}
\caption{Sample of detection tasks - 1}
\end{figure*}

\begin{figure*}[ht]
\begin{center}
  \includegraphics[width=1\linewidth]{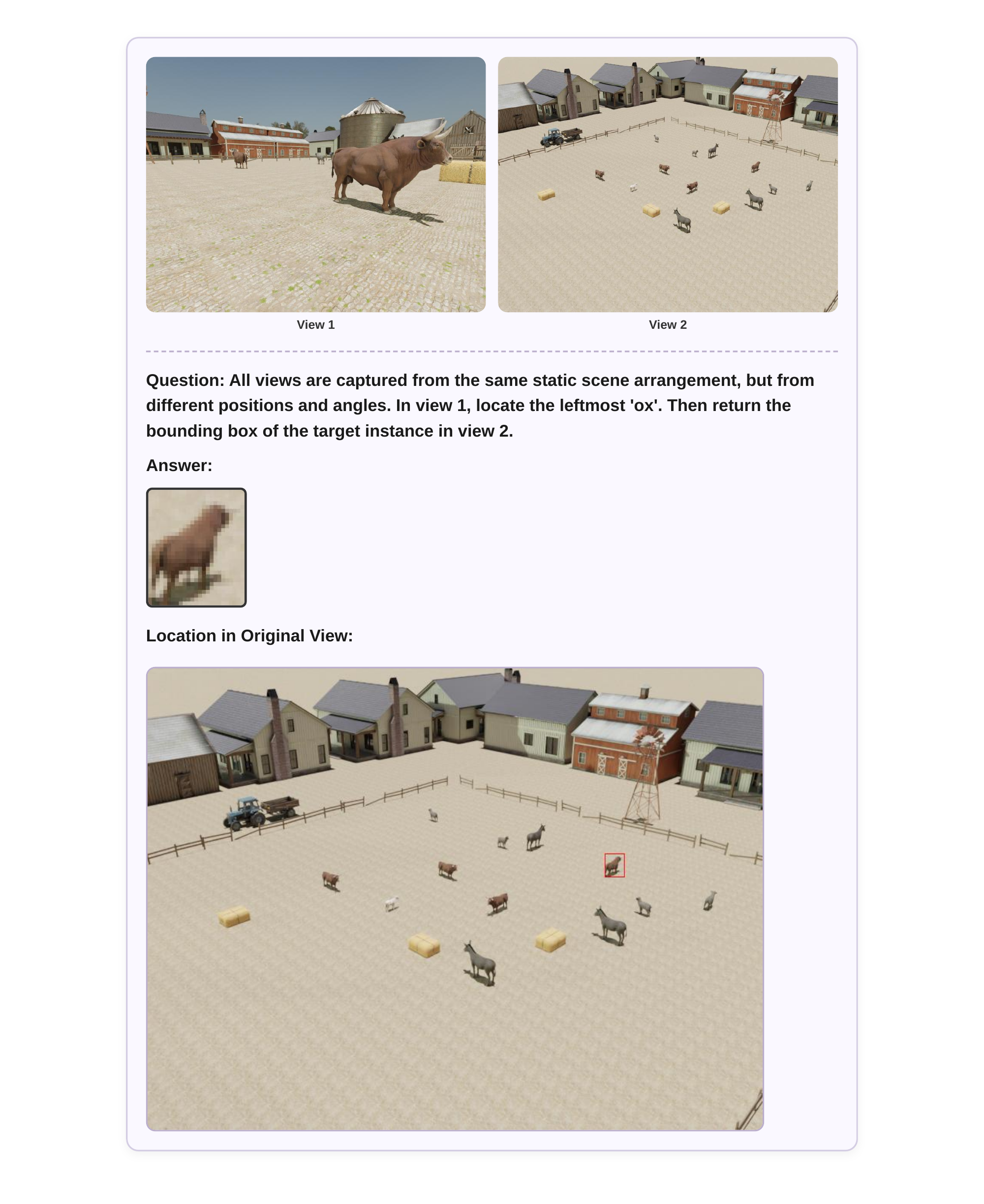} 
\end{center}
\vspace{-3em}
\caption{Sample of detection tasks - 2}
\end{figure*}

\begin{figure*}[ht]
\begin{center}
  \includegraphics[width=1\linewidth]{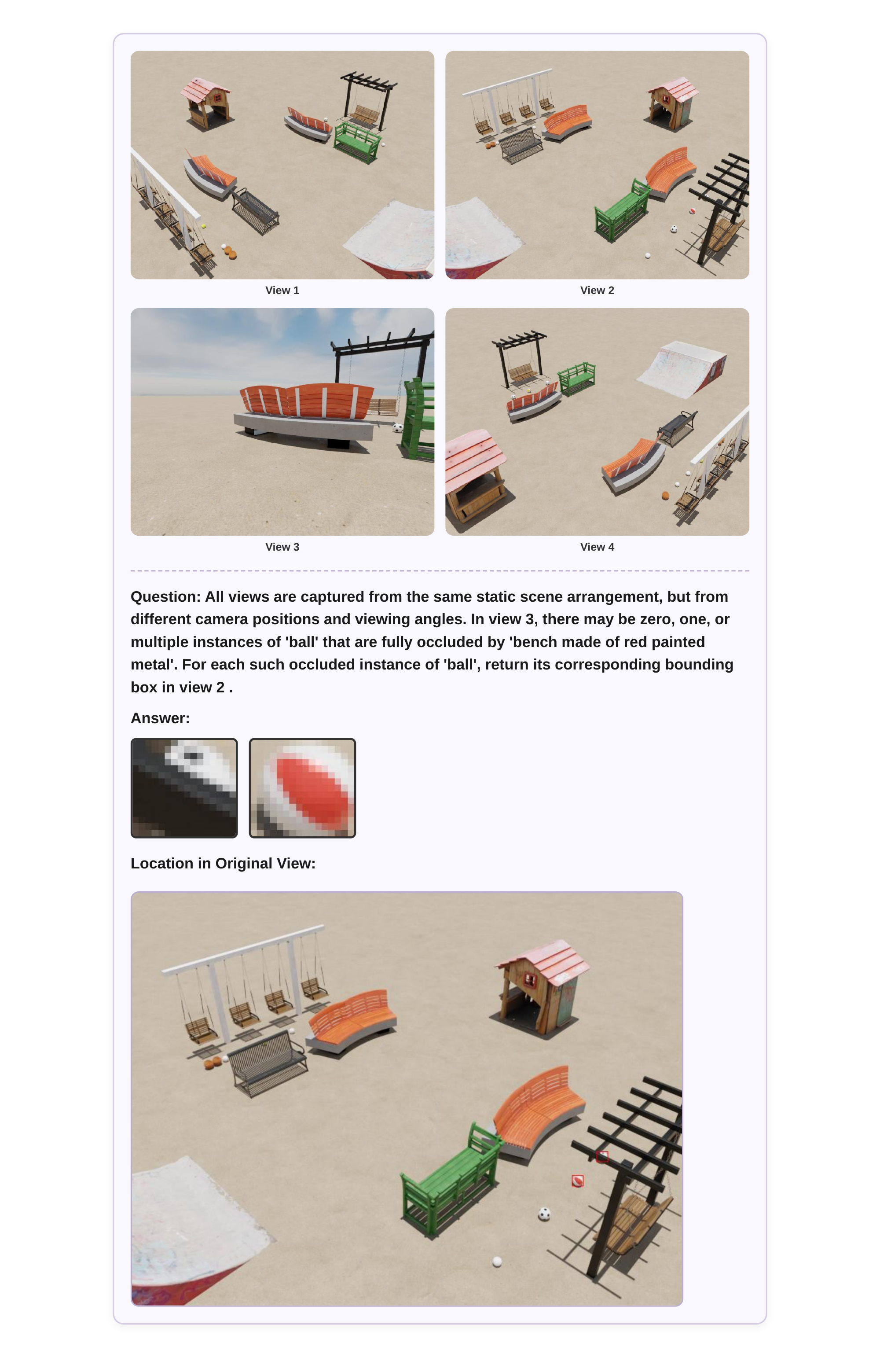} 
\end{center}
\vspace{-3em}
\caption{Sample of detection tasks - 3}
\end{figure*}

\begin{figure*}[ht]
\begin{center}
  \includegraphics[width=1\linewidth]{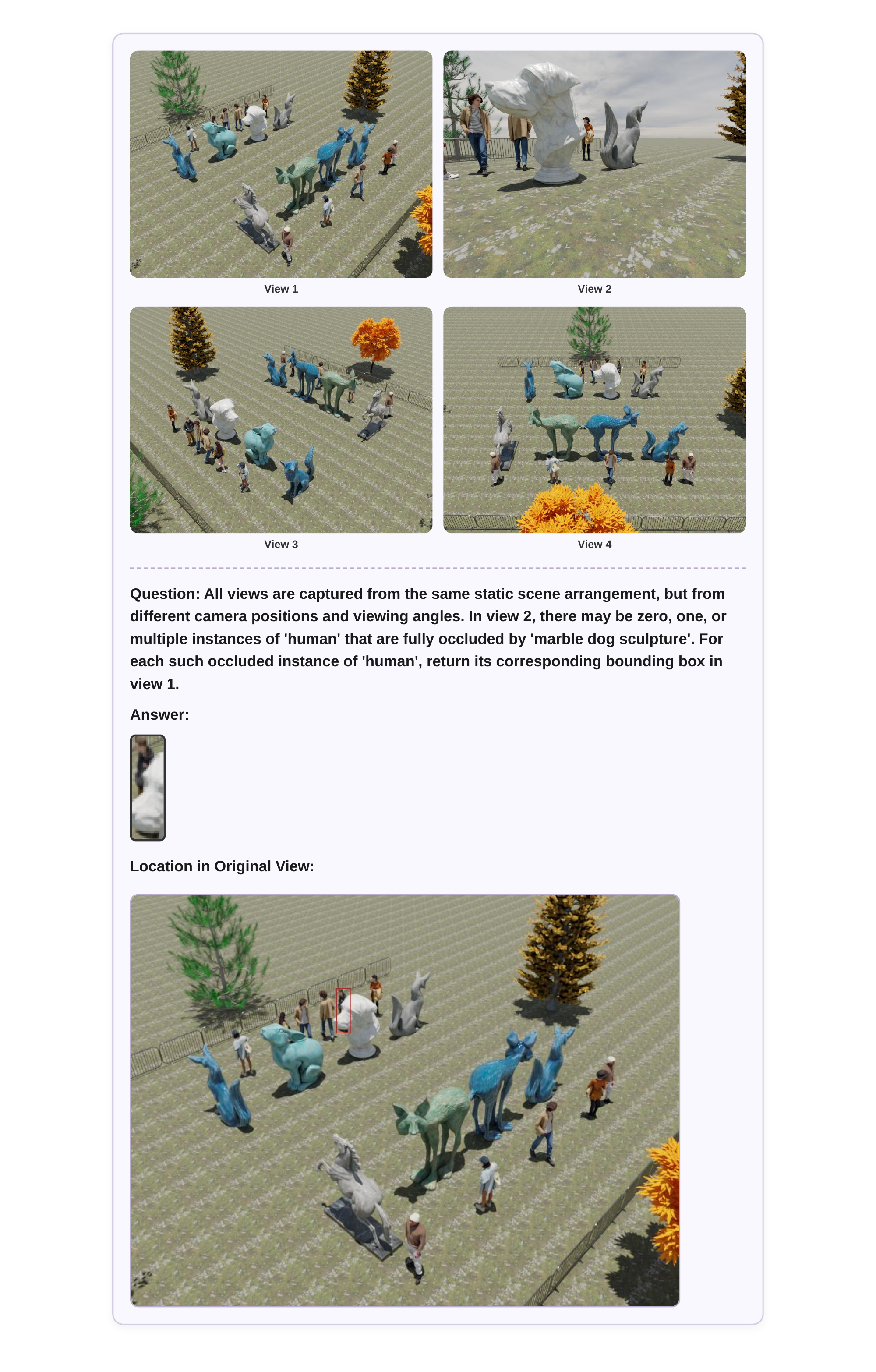} 
\end{center}
\vspace{-3em}
\caption{Sample of detection tasks - 4}
\end{figure*}

\begin{figure*}[ht]
\begin{center}
  \includegraphics[width=1\linewidth]{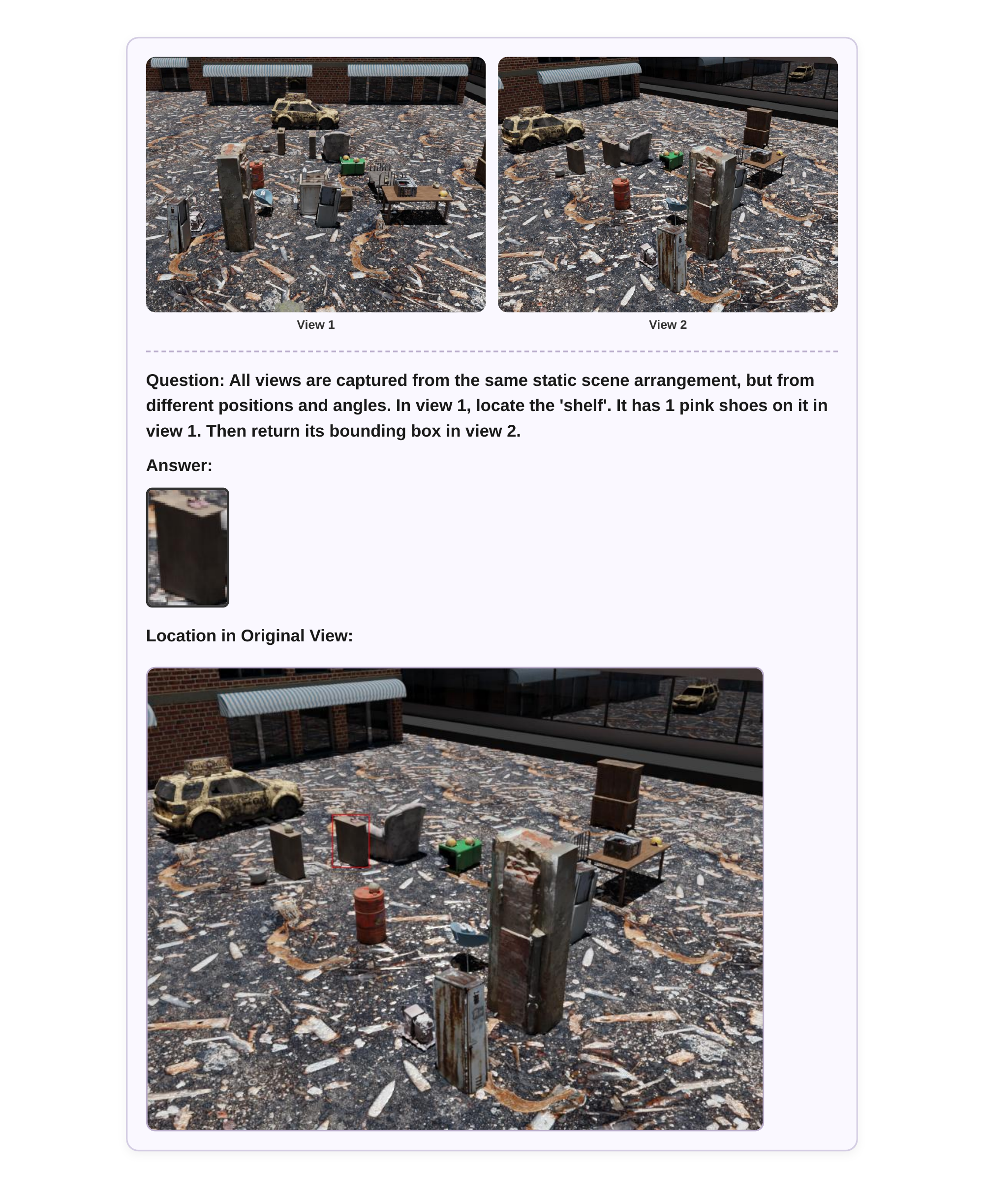} 
\end{center}
\vspace{-3em}
\caption{Sample of detection tasks - 5}
\end{figure*}

\begin{figure*}[ht]
\begin{center}
  \includegraphics[width=1\linewidth]{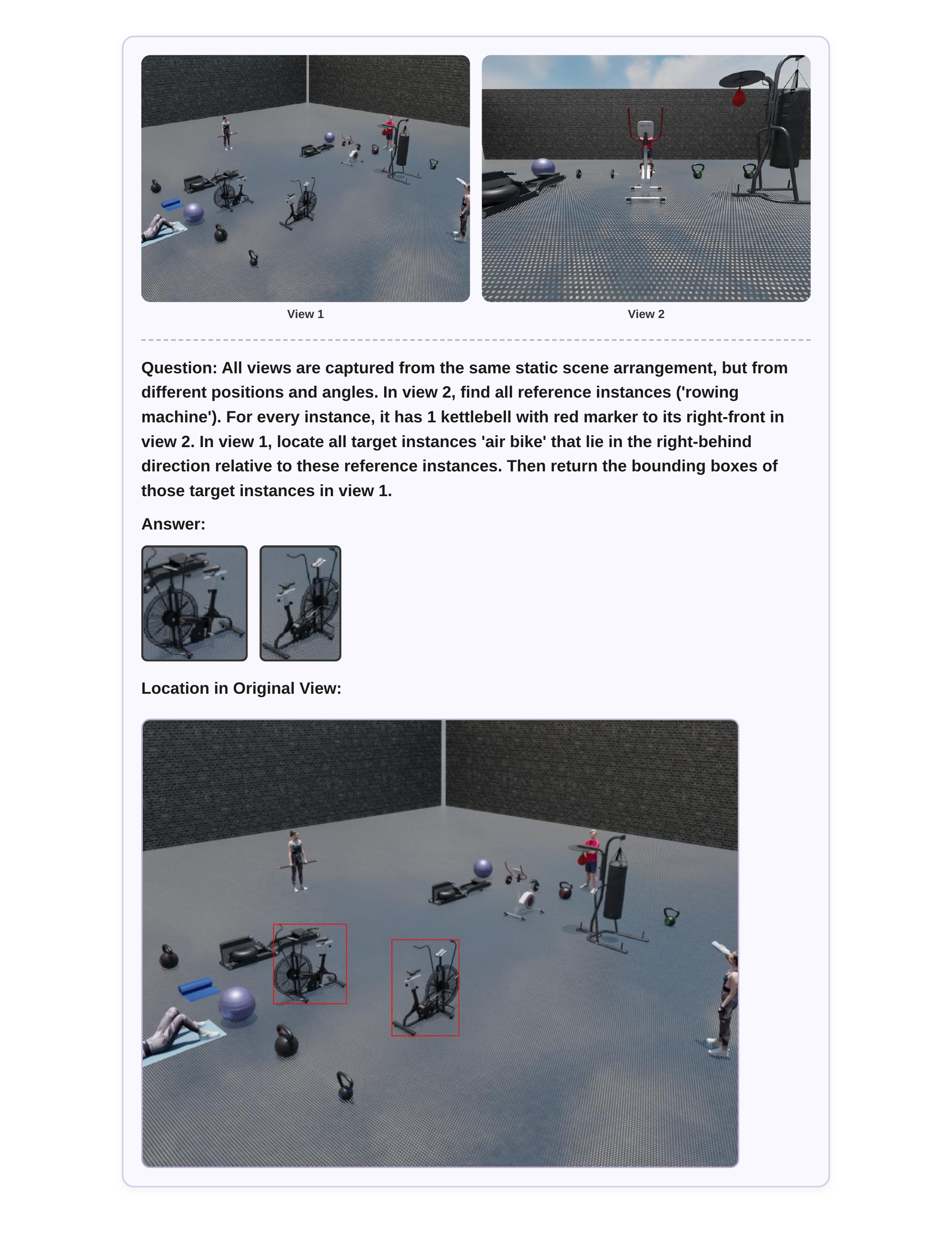} 
\end{center}
\vspace{-3em}
\caption{Sample of detection tasks - 6}
\end{figure*}

%
%
\clearpage
\bibliographystyle{splncs04}
\bibliography{main}
\end{document}